\documentclass{ieeeaccess}

\usepackage{xcolor,soul,framed} 

\colorlet{shadecolor}{yellow}
\usepackage[pdftex]{graphicx}
\graphicspath{{../pdf/}{../jpeg/}}
\DeclareGraphicsExtensions{.pdf,.jpeg,.png}

\usepackage[cmex10]{amsmath}
\usepackage{amsfonts}
\usepackage{textcomp}
\usepackage{pifont}
\usepackage{array}
\usepackage{mdwmath}
\usepackage{mdwtab}
\usepackage{eqparbox}
\usepackage{url}
\usepackage{algorithm}
\usepackage{algorithmic}
\usepackage{mathtools}
\usepackage[capitalise]{cleveref}
\usepackage{booktabs}
\usepackage{makecell}
\usepackage{subfig}
\usepackage{multirow}
\usepackage[numbers]{natbib}
\usepackage[figuresright]{rotating}

\allowdisplaybreaks

\newcommand{\cmark}{\ding{51}}
\newcommand{\xmark}{\ding{55}}

\usepackage{amsmath,amssymb,amsfonts}
\usepackage{algorithmic}
\usepackage{graphicx}
\usepackage{textcomp}
\usepackage{bm}
\makeatletter
\AtBeginDocument{\DeclareMathVersion{bold}
\SetSymbolFont{operators}{bold}{T1}{times}{b}{n}
\SetSymbolFont{NewLetters}{bold}{T1}{times}{b}{it}
\SetMathAlphabet{\mathrm}{bold}{T1}{times}{b}{n}
\SetMathAlphabet{\mathit}{bold}{T1}{times}{b}{it}
\SetMathAlphabet{\mathbf}{bold}{T1}{times}{b}{n}
\SetMathAlphabet{\mathtt}{bold}{OT1}{pcr}{b}{n}
\SetSymbolFont{symbols}{bold}{OMS}{cmsy}{b}{n}
\renewcommand\boldmath{\@nomath\boldmath\mathversion{bold}}}
\makeatother

\def\BibTeX{{\rm B\kern-.05em{\sc i\kern-.025em b}\kern-.08em
    T\kern-.1667em\lower.7ex\hbox{E}\kern-.125emX}}

\begin{document}

\history{Received 21 February 2025, accepted 12 March 2025, date of publication 27 March 2025.}
\doi{10.1109/ACCESS.2025.3555393}

    \title{Predict+Optimize Problem in Renewable Energy Scheduling.}
    
\author{
Christoph Bergmeir\authorrefmark{1,2,3}, 
Frits de Nijs\authorrefmark{3}, 
Evgenii	Genov\authorrefmark{10},
Abishek Sriramulu\authorrefmark{3},
Mahdi	Abolghasemi\authorrefmark{4},
Richard	Bean\authorrefmark{5},
John Betts\authorrefmark{3},
Quang	Bui\authorrefmark{3},
Nam Trong Dinh\authorrefmark{6},
Nils	Einecke\authorrefmark{7},
Rasul	Esmaeilbeigi\authorrefmark{8},
Scott	Ferraro\authorrefmark{9},
Priya	Galketiya\authorrefmark{9},
Robert	Glasgow\authorrefmark{9},
Rakshitha	Godahewa\authorrefmark{3},
Yanfei Kang\authorrefmark{11},
Steffen	Limmer\authorrefmark{7},
Luis	Magdalena\authorrefmark{12},
Pablo Montero-Manso\authorrefmark{13},
Daniel Peralta\authorrefmark{1,2},
Yogesh	Pipada Sunil Kumar\authorrefmark{6},
Alejandro Rosales-P\'erez\authorrefmark{14},
Julian	Ruddick\authorrefmark{10},
Akylas	Stratigakos\authorrefmark{15},
Peter	Stuckey\authorrefmark{3},
Guido Tack\authorrefmark{3},
Isaac Triguero\authorrefmark{1,2,16},
Rui	Yuan\authorrefmark{6}
}
\corresp{C. Bergmeir (e-mail: bergmeir@ugr.es) is the corresponding author. The first 4 authors are in order of contribution, the other authors are listed in alphabetical order. Affiliations are mostly affiliations at the time the work was performed.
}

\address[1]{ DaSCI Andalusian Institute in Data Science and Computational Intelligence, Granada, Spain.
}

\address[2]{ Department of Computer Science and Artificial Intelligence, University of Granada, Granada, Spain.
}

\address[3]{ Department of Data Science and Artificial Intelligence, Monash University, Melbourne, Australia.
}

\address[4]{ Queensland University of Technology, Brisbane, Australia.}

\address[5]{ Centre for Energy Data Innovation, School of Information Technology and Electrical Engineering, University of Queensland, Brisbane, Australia
}

\address[6]{ School of Electrical and Electronics Engineering, University of Adelaide, Adelaide, Australia
}

\address[7]{ Honda Research Institute Europe GmbH, 63073 Offenbach am Main, Germany
}

\address[8]{ School of Information Technology, Deakin University, Melbourne, Australia
}

\address[9]{ Building and Property Division, Monash University, Melbourne, Australia
}

\address[10]{ EVERGi, MOBI, Vrije Universiteit Brussel, Brussels, Belgium
}

\address[11]{ School of Economics and Management, Beihang University, Beijing, China
}

\address[12]{ E.T.S. Ingenieros Inform\'aticos, Universidad Polit\'ecnica de Madrid, Madrid 28660, Spain
}

\address[13]{ Disciple of Business Analytics, University of Sydney, Australia
}

\address[14]{ Department of Computer Science, Centro de Investigaci\'on en Matem\'aticas, Monterrey, 66629, Mexico
}

\address[15]{ Center for processes, renewable energy and energy systems (PERSEE), Mines Paris, PSL University, 06904 Sophia Antipolis, France
}

\address[16]{ The Optimisation and Learning (COL) Lab at the School of Computer Science, University of Nottingham, Nottingham NG8 1BB, United Kingdom
}

\tfootnote{Support roles for this competition received funding from ARENA as part of ARENA's Advancing Renewables Program. Christoph Bergmeir and Isaac Triguero are supported by María Zambrano (Senior) Fellowships that are funded by the Spanish Ministry of Universities and Next Generation funds from the European Union. This publication is supported by the Knowledge Generation Project PID2023-149128NB-I00, funded by the Ministry of Science, Innovation and Universities of Spain.
}

\markboth
{Bergmeir \headeretal: Predict+Optimize Problem in Renewable Energy}
{Bergmeir \headeretal: Predict+Optimize Problem in Renewable Energy}


\titlepgskip=-21pt

\begin{abstract}
Predict+Optimize frameworks integrate forecasting and optimization to address real-world challenges such as renewable energy scheduling, where variability and uncertainty are critical factors. This paper benchmarks solutions from the
IEEE-CIS Technical Challenge on Predict+Optimize for Renewable Energy Scheduling, focusing on forecasting renewable production and demand and optimizing energy cost. The competition attracted 49 participants in total. The top-ranked method employed stochastic optimization using LightGBM ensembles, and achieved at least a 2\% reduction in energy costs compared to
deterministic approaches, demonstrating that the most accurate point forecast does not necessarily guarantee the best performance in downstream optimization.
The published data and problem setting establish a benchmark for further research into integrated forecasting-optimization methods for energy systems, highlighting the importance of considering forecast uncertainty in optimization models to achieve cost-effective and reliable energy management. The novelty of this work lies in its comprehensive evaluation of Predict+Optimize methodologies applied to a real-world renewable energy scheduling problem, providing insights into the scalability, generalizability, and effectiveness of the proposed solutions. 
Potential applications extend beyond energy systems to any domain requiring integrated forecasting and optimization, such as supply chain management, transportation planning, and financial portfolio optimization.
\end{abstract}

\maketitle

\section{Introduction}

Optimization problems to be solved over an unknown future are at the core of
many complex real-world operations. For example, supply chains, inventories and
staffing rosters all need to be planned based on assumptions of future customer
demand. 

\begin{table*}[h!]
    \centering
    \caption{Summary of relevant studies and competitions in the field. Acronyms: NN - Neural Networks, CARTs - Classification and Regression Trees, RL - Reinforcement Learning, MIP - Mixed Integer Programming, EA - Evolutionary Algorithms, HCA - Hill Climbing Algorithm}
        \begin{tabular}{l|l|l|l|l|l|l|l}
        \hline
        \textbf{Ref} & \textbf{Type} & \makecell{\textbf{Optimization} \\ \textbf{Method}} & \makecell{\textbf{Forecasting} \\ \textbf{Evaluation}} & \makecell{\textbf{Demand} \\ \textbf{Response}} & \makecell{\textbf{Timetable} \\ \textbf{Optimization}} & \makecell{\textbf{Storage} \\ \textbf{Optimization}} & \makecell{\textbf{Open} \\ \textbf{Data}} \\
        \hline
        \citet{donti2017task} & methodology & custom loss NN & \cmark & \xmark & \xmark & \cmark & \xmark \\
        \citet{stratigakos2022prescriptive} & methodology & custom loss CARTs & \xmark & \xmark & \xmark & \cmark & \cmark \\
        \citet{jonban2024reinforcement} & methodology & RL & \xmark & \xmark & \xmark & \cmark & \xmark \\
        \citet{genov2024predict} & methodology & stochastic MIP & \cmark & \xmark & \xmark & \cmark & \cmark \\
        \citet{salari2024model} & methodology & RL & \xmark & \cmark & \xmark & \cmark & \xmark \\
        \citet{Han20215294} & methodology & custom loss NN & \cmark & \cmark & \xmark & \cmark & \xmark \\
        \citet{vazquez2020citylearn} & competition & RL & \xmark & \cmark & \xmark & \cmark & \cmark \\
        \citet{nagy2021citylearn} & competition & RL & \xmark & \cmark & \xmark & \cmark & \cmark \\
        \citet{nweye2023citylearn} & competition & MIP,RL & \xmark & \cmark & \xmark & \cmark & \cmark \\
        \citet{van2017multi} & competition & HCA & \cmark & \xmark & \xmark & \cmark & \cmark \\
        \hline
        Current Study & competition & MIP,EA & \cmark & \cmark & \cmark & \cmark & \cmark \\
        \citet{Stratigakos2021} & participant & MIP &  &  &  &  &  \\
        \citet{ruddick2022evolutionary} & participant & EA,MIP &  &  &  &  &  \\
        \citet{abolghasemi2021state} & participant & MIP &  &  &  &  &  \\
        \citet{bean2022methodology} & participant & MIP &  &  &  &  &  \\
        \citet{limmer2022efficient} & participant & MIP &  &  &  &  &  \\
        \end{tabular}
    \label{tab:competitions}
\end{table*}

This type of optimization will also play a vital role in the global transition
to reduce CO2 emissions. Renewable energy production is characterized by
variability over time, and the inability to readily vary production based on
demand. Therefore, demand needs to be scheduled to make the best use of supply
where possible, with energy storage systems such as batteries scheduled
optimally to make up the shortfall, all based on unknown future production and
demand. The common approach to solve these problems is to forecast the future,
and use this as the ``true" input for the optimization. Although this is
expedient, it pays little regard to the uncertainty of the forecast. One way to
address uncertainty is to use robust optimization \cite{ben2009robust} or
stochastic optimization \cite{juan2015review}, with probabilistic forecasts as
inputs instead of point forecasts. Some applications along these lines are
presented by \citet{dehghani2021proactive} for trans-shipment and
\citet{jung2004simulation} for supply chain management.

Forecasting is often a precursor to optimization, which can minimize costs, maximize renewable energy use, or ensure energy system stability. Accurate forecasting is crucial because it provides the necessary inputs for the optimization models.
Without accurate forecasts, the optimization process may rely on incorrect or incomplete information, leading to suboptimal decisions. For instance, in the context of renewable energy scheduling, inaccurate forecasts of energy demand or solar production can result in either overestimating or underestimating the required energy storage, leading to increased costs or energy shortages.

More recently, researchers have started to address these types of problems more holistically in an emerging research field known as \emph{Predict+Optimize},
where forecasting and optimization are not treated as isolated tasks, but their
interaction is taken into account. Using this approach, forecasts are chosen or
evaluated through their contribution to the actual downstream cost of the
optimization problem, in preference to arbitrary measures of forecast quality,
such as Mean Absolute Error (MAE), Root Mean Square Error (RMSE), or Continuous
Ranked Probability Score (CRPS). \citet{ijcai2021p610} give an overview of
methods of this type. The most complete review of studies on integrated
forecasting and optimization is found in \citet{mandi2024decision}. The
integration of forecasting and optimization is also studied on the operational level, including studies on finding optimal prediction horizons \cite{prat2024long}, policy and forecast revision frequency
\cite{genov2024predict}, and decision-making under uncertainty using robust
optimization with distributional probabilistic forecasts \cite{stratigakos2022prescriptive}.
Another scope of integration is direct optimization of the forecast model with a decision-focused loss function. \citet{mandi2024decision} categorize integration methods as
gradient-based or gradient-free. Gradient-based methods use differentiable optimization mappings to backpropagate through optimization \cite{amos2017optnet,kotary2023backpropagation, donti2017task, gao2021online},
while gradient-free methods optimize directly for minimal regret \cite{elmachtoub2022smart,demirovic2020dynamic, mandi2020smart}.

\citet{elmachtoub2020decision}
develop a specialized algorithm to build decision trees directly for the true
optimization target, and \citet{elmachtoub2022smart} develop a differentiable
surrogate for the true optimization target. \citet{mandi2020smart} extend this
framework to discrete optimization and \citet{demirovic2020dynamic} develop an
alternative solution for dynamic programming. Other approaches aim to build
end-to-end systems where no intermediary forecasting is needed. For example,
\citet{donti2017task} build an end-to-end model in the form of a neural network
that optimizes for the loss of a stochastic optimization problem, and
\citet{gao2021online} build an end-to-end system using imitation learning for
scheduling of a microgrid.

A lot of research in these areas has focused on energy production and
consumption. In addition to classical work on forecasting, e.g., energy demand
\cite{Dudek20222879}, renewable power production
\cite{liao2021windgmmn,zhang2022very,liu2021bidirectional}, or energy price
\cite{Hu2020612}, there is an increasing body of work on machine learning
systems that integrate prediction and optimization via customized loss
functions, e.g., using gradient boosted regression trees (GBRTs) and neural
networks (NNs)
\cite{Li20182508,Han20215294,Stratigakos2021,Zhang2022,stratigakos2022prescriptive}.
Other research has used meta-optimization, in which an outer optimization loop
is added to a Predict+Optimize pipeline, such as in \citet{Carriere20196933},
and with methods to keep computational cost manageable, such as Lagrangian
relaxation \cite{Chen20223104} or simplified linear programs
\cite{Munoz20203753}. Because optimization and forecasting are both difficult
problems in their own right, the combined complexity of Predict+Optimize models
in the literature may not be applicable to real-world problems. It may be that
the combined problem requires simply specified problem instances, or that
computation time is prohibitively long. Realistic problems require both a
real-world data set and complex optimization. In this regard, there is a lack of
problems from which to systematically determine the state of the art in this
field of research. 

Competitions are an effective way to establish standard benchmark problems. With
a monetary prize as an incentive, participants are motivated to deliver the
best-performing solutions while pushing the boundaries of innovation to gain an
edge over others. In the context of energy management, for example there is ``The Citylearn Challenge'', a competition series that has been hosted in 2020
\cite{vazquez2020citylearn}, 2021 \cite{nagy2021citylearn} and 2022
\cite{nweye2023citylearn}. These competitions focus on optimizing building
energy use by scheduling flexible loads and batteries. While the initial
editions were specifically designed for reinforcement learning (RL), the 2022
edition also accommodates more general optimization methods, including Model
Predictive Control (MPC) configurations with Mixed Integer Programming (MIP) and
Mixed Integer Linear Programming (MILP) solvers. These classical optimization
methods demonstrated superior performance compared to RL in the competition.
However, RL contributed valuable diversity and adaptability through its capacity
for online learning. Notably, the top solution leveraged an ensemble of MIP and
RL, effectively combining the robust optimization capabilities of MIP with the
adaptive learning strengths of RL, leading to further performance improvements.
Although look-ahead predictions and their uncertainty quantification were
typically crucial for top-performing solutions in these competitions, the
Citylearn challenges did not specifically analyze the role of forecasting to the
downstream optimization. For the combined evaluation and benchmarking of both
subproblems, only one competition in this area is known to us, the ``ICON
Challenge on Forecasting and Scheduling,'' hosted in 2016. This challenge
required a single time series (energy price) to be forecast, for the subsequent
scheduling of server jobs to minimize energy cost. With a relatively simple
prediction problem and a difficult optimization problem, this challenge leaned
heavily towards optimization. The competition winner \cite{van2017multi}
implemented heuristics for generating an initial solution, which was then
improved using a hill climbing algorithm. 

Inspired by the ICON Challenge, we organized the ``IEEE-CIS Technical Challenge
on Predict+Optimize for Renewable Energy Scheduling," \cite{Bergmeir2021IEEE},
as part of a series of yearly Technical Challenges hosted and sponsored by the
IEEE Computational Intelligence Society. The goal of this challenge was to
provide a relevant real-world dataset and optimization benchmark problem along
with strong baseline solutions from which to establish a state of the art in the
area for the research community. We hope that this will enable more standardized
and streamlined evaluation of future research in the field. A particular aim of
the competition was to balance the requirements of the problem so that the
competition could not be won by focusing on either forecasting or optimization
alone. The comparison to studies related to Predict+Optimize problems in energy
management, as well as relevant competitions, is provided in
\cref{tab:competitions}. 

The table highlights that the current study is the report of a competition that
is unique in its focus on the Predict+Optimize problem, presents a balanced and
challenging problem, provides an open-access benchmark for forecasting and
optimization for further research, and is open to a wide range of solution
methods. Furthermore, the discussed problem includes a timetable scheduling
component, which is not present in the other competitions. The objective of this
paper is to provide an overview of state-of-the-art solutions for
Predict+Optimize problems in renewable energy scheduling. We assume that the
competition establishes a meaningful state of the art not only by analyzing the
solutions with the help of a scientific committee, but also through the
available prize money. If solutions were not state of the art, somebody else
could have come in to win the competition and the prize money. By analyzing
these solutions, this study aims to support the development of more reliable and
cost-effective renewable energy strategies while advancing methodological
research in this domain.

This paper reviews the solution methods proposed by participants in the competition. The complete reports of these solutions are available in Appendix A with the supplementary material, as well as in the proceedings from this competition, also shown in \cref{tab:competitions}. This paper reflects on
the competition setup, the solutions submitted, and the final rankings. The
results are analyzed with respect to the performance, common themes, and best
practices. The key contributions are as follows:
\begin{enumerate}
    \item A comprehensive evaluation of Predict+Optimize methodologies applied to a real-world renewable energy scheduling problem.
    \item The establishment of a benchmark for future research, including insights into the scalability, generalizability, and effectiveness of the proposed solutions.
    \item A synthesis of best practices and innovative strategies demonstrated by the top-performing teams.
\end{enumerate}

The remainder of the paper is structured as follows.
Section~\ref{sec:comp_setup} presents the competition setup.
Section~\ref{sec:sol} discusses the submitted solutions, presents the final
rankings, and gives an overall summary of the results.
Section~\ref{sec:det_desc} describes the best-performing solutions, Section~\ref{sec:discussion} discusses the results and
Section~\ref{sec:conc} concludes the paper.

\section{Competition setup}
\label{sec:comp_setup}

\begin{figure}[h]
    \centering
    \includegraphics[width=3.4in]{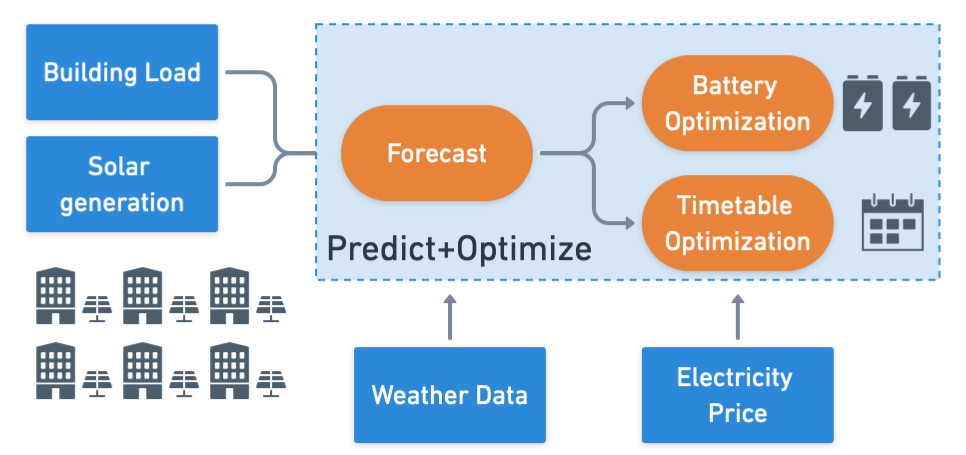}
    \caption{Implied data flow in the problem setting.}
    \label{fig:datachart}
\end{figure}

\begin{figure}[h]
\centering
\includegraphics[width=3.4in]{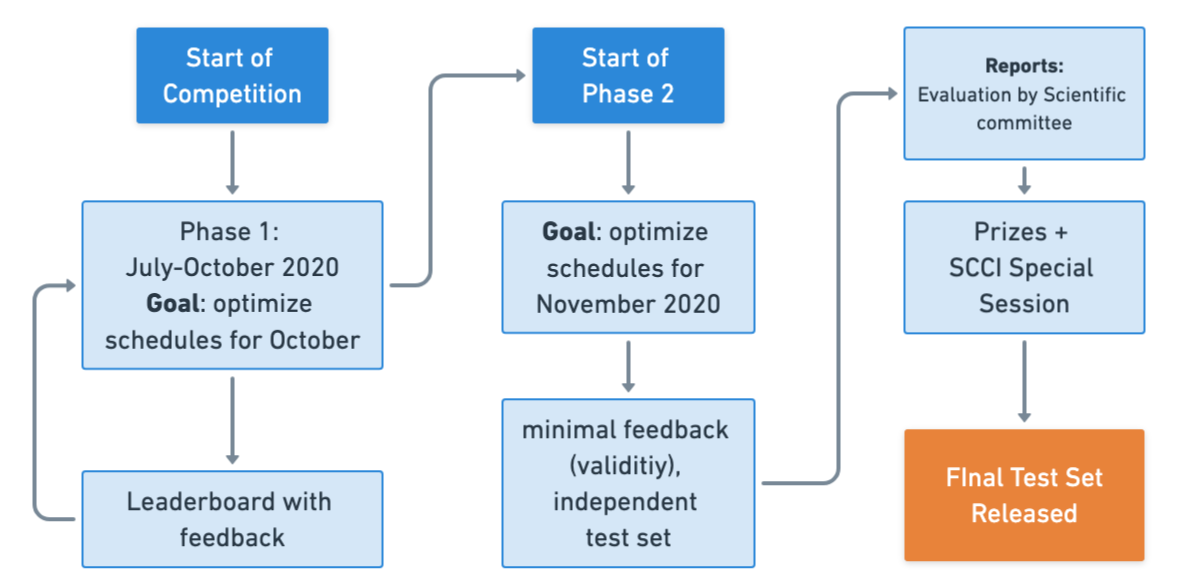}
\caption{Competition setup.}
\label{fig:Flowchart_setup}
\end{figure}
\cref{fig:datachart} illustrates the data flow associated with the problem
formulation. Forecasting and optimization can be viewed as either subproblems or
components of an integrated solution. The problem originates from the Monash Net
Zero Initiative, which involves a campus-wide microgrid equipped with rooftop
solar photovoltaic installations and a battery for energy storage. This system
operates across the entire university campus, including buildings, grounds, and
electric vehicles (EVs), with the goal of achieving net-zero emissions by 2030.
In particular, it aims to: 1) maximize self-consumption of electricity, 2)
participate in energy demand response programs, and 3) keep track of electricity
price and yearly peak load tariffs, to manage costs.

From a technical point of view, the data provided presents an interesting time
series prediction problem. The demand and production data has complex
seasonality; external data (weather, electricity price) are factors in the
problem. There is also the opportunity for cross-learning between time series
for the energy demand and solar production problems. From an optimization point
of view, the uncertainty in the inputs presents a mismatch between the forecast
(production and demand) and that which actually eventuates. This needs to be
addressed, along with various constraints to achieve a competitive solution. The
goal of the optimization is to develop a battery charge and discharge schedule
along with a lecture theatre use schedule that results in the lowest energy
cost. Battery use is constrained by capacity. Lecture theatre use is determined
by the university timetable, with some activities being regular, and others
one-off. 

The competition setup aimed to be as close to a typical real-world situation as
possible. However, some adjustments needed to be made, due to the nature of the
competition. In real-life, battery scheduling would typically be performed using
historic data such as: building demand, solar power production, weather or
specialized solar forecasts, and electricity price forecasts (from external
providers) as input variables. Based on this, the battery schedule would then be
optimized for approximately 1-3 days in advance, re-running the optimizer
periodically (e.g., every 15 minutes). In real-life, lecture times and locations
would be planned well in advance of the academic year, and without regard to the
power schedule. Building energy use would ideally be comprised of building base
load, along with the energy use from scheduled demand.

This setup was designed to establish a consistent benchmark for participants
while addressing key challenges, particularly the prevention of data leaks. To
ensure fairness, the competition assumed perfect knowledge of weather and energy
prices, eliminating uncertainties associated with longer planning horizons. This
allowed for meaningful comparisons with real-world short-term scheduling
approaches, such as day-ahead planning with continuous updates. The competition
structure was carefully chosen to strike an optimal balance between realism and
feasibility, given the offline nature of the challenge.

\begin{figure*}
    \centering
    \includegraphics[width=\textwidth]{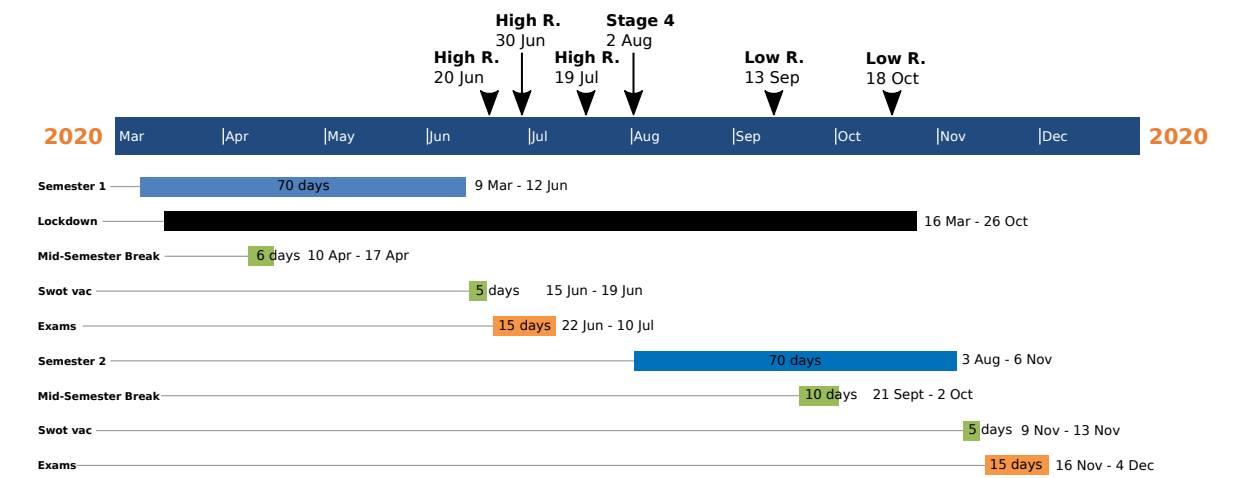}
    \caption{Timeline of Melbourne lockdown measures in 2020 due to the COVID-19 pandemic.}
    \label{fig:covline}
\end{figure*}

A crucial design decision was whether to conceal the exact time and location of
the energy usage data—preventing participants from cross-referencing publicly
available weather and market data—or to assume this information was known. The
latter approach was adopted, meaning the competition effectively assumed access
to perfect forecasts for weather and energy prices. Additionally, the dataset’s
source—the Monash Clayton campus—was disclosed, along with guidance on how
participants could retrieve relevant external data. Due to Melbourne’s COVID-19
lockdowns in 2020, on-campus lectures were suspended, significantly reducing
activity levels. This allowed participants to make reasonable approximations of
baseline building loads. A timeline of events affecting campus operations is
provided in Figure~\ref{fig:covline}, while Table~\ref{tab:covtable} summarizes
key events and the estimated campus occupancy levels during the period.

\begin{table}[htb]
\centering
\includegraphics[width=3.4in]{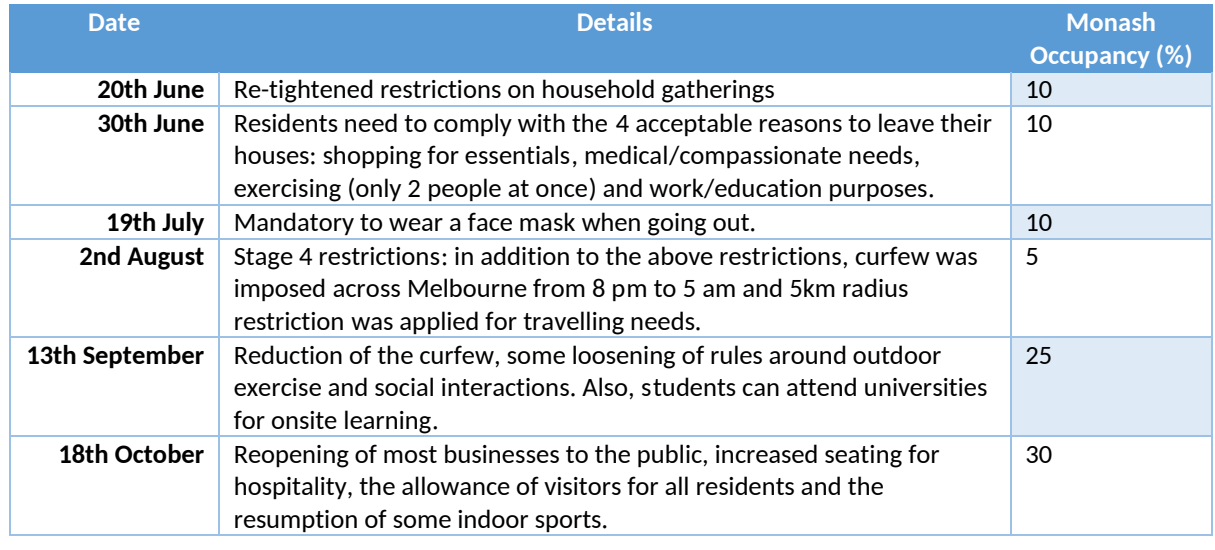}
\caption{Table of events leading to increase or reduction of lockdown restrictions.}
\label{tab:covtable}
\end{table}

\cref{fig:Flowchart_setup} shows the setup of the competition organized in two
phases. Phase~1 ran for 3 months, from July to October 2021. Phase~2 ran for
approximately 3 weeks during October 2021. The goal of Phase~1 was to optimally
schedule batteries and timetabled activities (lectures) for the month of October
2020. Participants could submit forecasts and/or optimal schedules to a
leaderboard during this phase. These were then evaluated, with the results
visible to all participants. Participants were also provided with na\"ive sample
submissions for both forecasting and optimization. In particular, the
forecasting submission provided was a forecast of constant zeros and the
optimization solution provided was a greedy scheduling solution.

During Phase~2 of the competition, data for October 2020 was released to the
participants, who were  asked to perform the same forecasting and optimization
exercise as Phase~1, but for November 2020. Several problems in the competition
setup were addressed at this time. Notably, time zones for forecasting and
optimization were aligned, and we ensured that no large outliers equivalent to
those identified in the Phase~1 test data set (which made forecasting less
important) were present in the Phase~2 test set.

During this phase, only minimal feedback was provided to the participants about
the quality of their submissions in the form of whether the solution was valid,
and whether it was better, equal, or worse than the sample submission. 

Phase~2 of the competition determined the winners and prizes awarded. The
majority of prizes (USD \$18k from a total of USD \$20k) were awarded based on
optimal energy cost. Teams could choose any methodology for optimization. This
included the freedom not to perform forecasting if this was deemed unnecessary.
However, a separate forecasting prize of USD \$2k was awarded to the best
forecasting solution, to encourage the participants to consider forecasting as
part of their solution, and to promote a competition that integrates forecasting
and optimization.

The competition was set up in line with best practices from the research
literature and competition platforms such as Kaggle~\cite{kaggle}. In
particular, \citet{athanasopoulos2011value} argue that feedback in competitions
leads to better outcomes, which is why Phase~1 of the competition presented
results transparently. This enabled participants to gain a deep understanding of
the problem, and also gave the organizers an opportunity to identify and address
problems in the competition setup. The independent test set and minimal feedback
in Phase~2 ensured that participants had no means to overfit their energy
forecasts, but still had a mechanism to ensure their solution was valid.

Unlike many competitions where a single solution determines its ranking, a
scientific committee consisting of 8 scholars was assembled to rank the
submissions according to certain criteria (see Section~\ref{sec:sc_eval}), based
on a 4-page report of the methodology submitted by each of the shortlisted
teams. This additional score was then combined with the optimization
scores to determine a final score. The aim of this exercise was to ensure the
scientific rigor and benefit to the research community of the winning solutions
by promoting those with more general applicability in practice, over those that
were very tailored to the competition data and the evaluation metrics. 

Once winners were determined and prizes awarded, the final test set of November
2020 was released, so that the solutions where participants published their code
could be reproduced.

\subsection{Data description}

The following energy consumption, solar production and weather data was made available to participants from the competition web page \cite{Bergmeir2021IEEE}, where it continues to be publicly available.

\begin{figure*}[htb]
    \centering
    \subfloat[Weather Data: Temperature and Surface Radiation]{
        \includegraphics[width=0.48\textwidth]{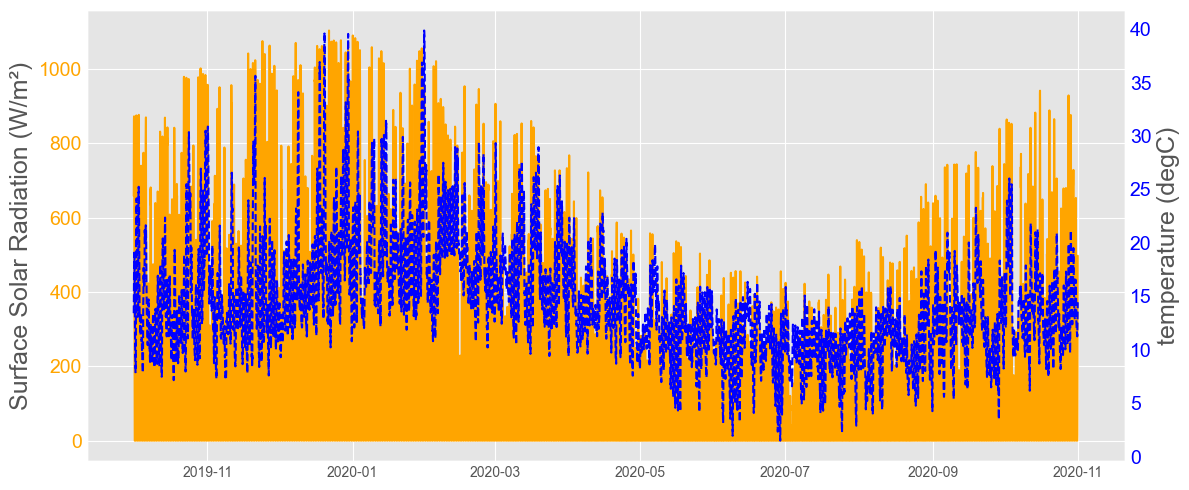}
    }
    \hfill
    \subfloat[Electricity Price for October 2020]{
        \includegraphics[width=0.48\textwidth]{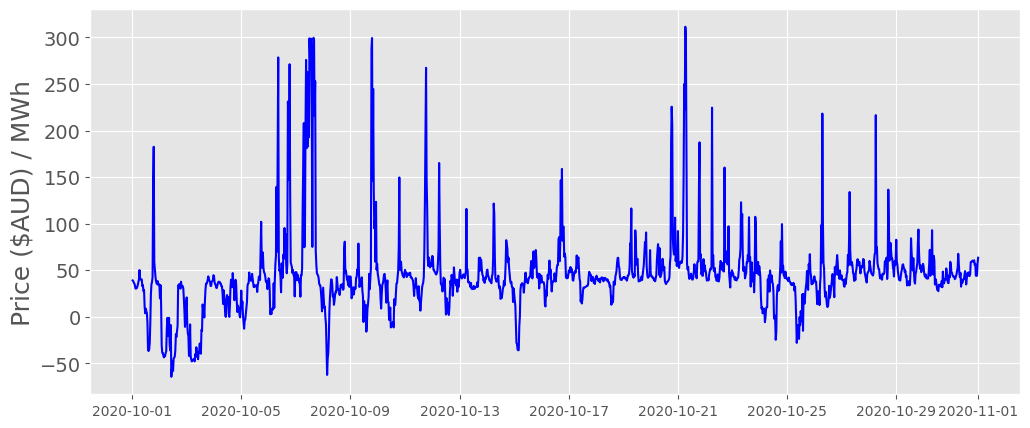}
    }
    \caption{(a) Weather data of temperature and surface radiation from ERA5 for the location of interest. (b) Time series of electricity price for October and November 2020 (validation and test period).}
    \label{fig:price_weather}
\end{figure*}

\begin{figure*}
\centering
\includegraphics[width=\textwidth]{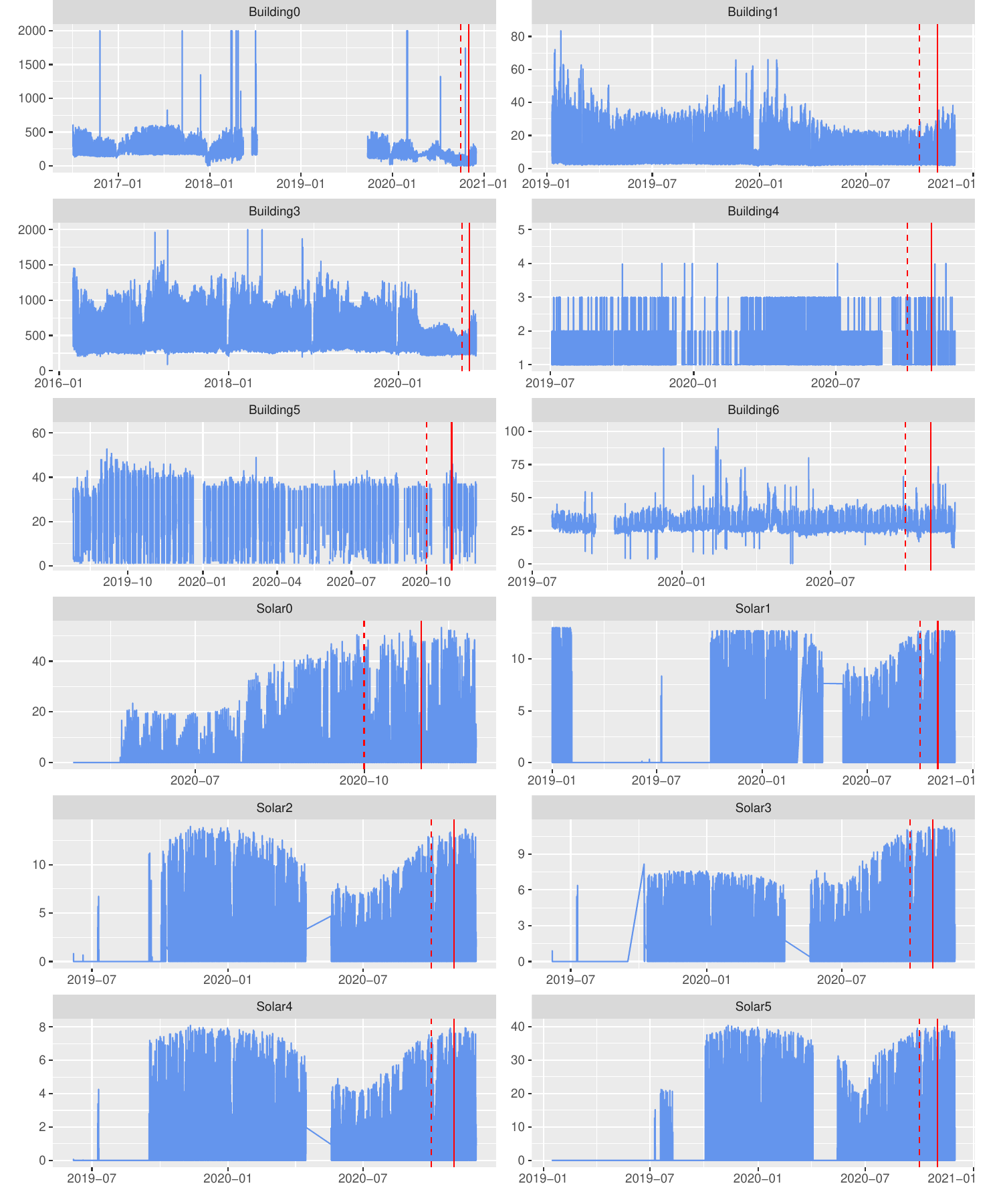}
\caption{Input series of building load and solar power
production from the Monash Clayton campus. All values are in $kW$. Building~0
and Building~3 have large outliers that have been capped at 2000. The dashed
lines indicate the start of the Phase~1 test data in the competition, the solid
lines indicate the start of the Phase~2 test data (i.e., data for October and
November 2020, respectively). } 
\label{fig:inputseries}
\end{figure*}

\begin{itemize}
    \item \textbf{Energy consumption data} recorded at 15-minute intervals was
    obtained from 6 buildings on the Monash Clayton campus over varying time
    periods, up to September 2020 (for Phase~1) and October 2020 (for Phase~2).
    Time series of about 5 years, commencing in 2016 were obtained from
    Buildings 0 and 3, whereas shorter time series of about one year were
    obtained from the other buildings. The dataset doesn't contain a building
    numbered 2 as the data for this building was scarce and the decision was
    made to exclude this building before the competition started.

    \item \textbf{Solar production data} from 6 rooftop solar installations on the Monash Clayton campus was recorded at 15-minute intervals over approximately one year until September/October 2020 (for Phases~1 and 2 of the competition, respectively). These data (in $kW$) are also shown in Figure~\ref{fig:inputseries}. One participant noted that the data for Solar~1 seem to be cumulative data for some parts of the
    series.
    \item \textbf{Weather data (ERA5)} was generously provided by
    Oikolab~\cite{oikolab}. It contains hourly measurements of temperature
    ($^{\circ}C$), dewpoint temperature ($^{\circ}C$), wind speed ($m/s$), mean
    sea level pressure (Pa), relative humidity ($0-1$), surface solar radiation
    ($W/m^2$), surface thermal radiation ($W/m^2$), and total cloud cover
    ($0-1$), from 2010 to 2021. The series for temperature data and surface
    radiation are shown in Figure~\ref{fig:price_weather}(a). The
    temperature and surface radiation data show clear daily and seasonal
    patterns typical for the Southern Hemisphere, with higher values during the
    summer months and lower during the winter months. This data is crucial for
    predicting energy demand and solar power production.
\end{itemize}

\noindent Participants were also encouraged to use the following data from external sources:

\begin{itemize}

    \item  \textbf{Weather data (BOM)} from the Australian Bureau of Meteorology
    (BOM) included the daily minimum temperature ($^{\circ}C$), maximum
    temperature ($^{\circ}C$), rainfall ($mm$) and solar exposure ($MJm^{-2}$)
    at three weather stations near the Monash Clayton campus: Olympic Park,
    Moorabbin Airport and Oakleigh (Metropolitan Golf Club)~\cite{bom-climate}.
    Each data series commenced on the 1st of January 2016 and concluded at the
    date of download by participant. 
    \item \textbf{Electricity price data} from the Australian Energy Market
    Operator (AEMO) consisted of half-hourly electricity price and demand data
    at the state level ~\cite{aemo-data}. For Phase~1 of the competition, the relevant data was Victoria during
    October 2020, available from \cite{aemo-vic-data}. The price time series data are shown
    in Figure~\ref{fig:price_weather}(b) for the period of Phase 1. \footnote{Though the intended use from this data source was the price
    data, some participants also used the demand data that they found helpful.
    The competition policy stated to gain permission from the organizers for any
    external datasets. However, as the demand data was (unintentionally)
    provided by the organizers, it was a grey area so that teams using the
    dataset did not request permission and in consequence not all teams were
    aware of the demand data, and that it could potentially be used.}
    
\end{itemize}

\subsection{Forecasting}

Forecasting was optional in the competition, but encouraged through a small
prize for the most accurate forecast. During Phase~1 of the competition,
participants were expected to predict the power demand of the 6 buildings, and
the power production of the 6 arrays of solar panels over 15-minute intervals
for each day in October 2020. This amounted to 2976 15-minute forecasts in
total. At the end of Phase~1, the actual energy demand of each building and
power production of each array of solar panels was released. For Phase~2,
participants were then expected to provide the 15-minute energy
demand/production forecasts for the same buildings and solar panels over the 30
days of November 2020.

\subsection{Optimization}

The optimization problem was to create a timetable for a set of activities over
the coming month, across the set of six buildings, with the objective of
minimizing the total electricity cost. Power was provided at no cost by 6 sets
of solar panels as well as being available at market price from the energy grid.
Batteries were available, enabling the storage of excess solar energy for later
use, or charging from the grid during periods of low cost.

Participants were given 5 large, and 5 small problem instances. Each instance
consisted of a number of activities to be scheduled over the month, including
precedence requirements, and whether they were one-off or recurring. Each
building was specified by a number of large and small rooms available. Solar
production attributable to each building varied between scenarios by assigning
one of the six solar time series (see Figure~\ref{fig:inputseries}) to the
building in each instance. Activities were specified by the number of rooms
required, whether these rooms were large or small, the duration of the activity,
energy load of that activity, and whether it was recurrent or one-off. The
battery specifications were matched to be close to the actual scale and
performance of the two batteries currently installed in the Monash microgrid.
Each battery was characterized by its capacity, maximum charge/discharge power,
and roundtrip efficiency. The specifications for the two batteries used in all
problem settings are shown in Table~\ref{tab:battery_specs}.

\begin{table}[htbp]
\centering
\caption{Battery Specifications}
\begin{tabular}{|c|c|c|c|}
\hline
\textbf{Battery} & \textbf{Capacity} & \textbf{Max Power} & \textbf{Roundtrip} \\
 & \textbf{(kWh)} & \textbf{(kW)} & \textbf{Efficiency} \\
\hline
Battery 1 & 150 & 75 & 0.85 \\
Battery 2 & 420 & 60 & 0.60 \\
\hline
\end{tabular}
\label{tab:battery_specs}
\end{table}

The objective consists of three parts: 1) the wholesale energy cost of all
energy imported, 2) a peak load demand charge, and 3) the additional profit of
scheduled once-off activities. Given the cumulative power draw~$\overline{P}_t$
at time $t$ in kW, and the wholesale energy price~$e_t$ in \$/MWh (a time series
provided to the participants), power usage is converted to energy consumption by
assuming a constant load during each 15-minute time step. The demand charge is
fixed at \textdollar5 per MW, regardless of when the peak load occurs during the
month. Finally, the value of each scheduled ($d_i = 1$) once-off activity~$i$ is
earned, minus any penalty for scheduling outside office hours if applicable
($o_i = 1$). The combined objective function of the optimization is given by:
\begin{align}
O = {} & \sum_{t} \left( \frac{0.25 \overline{P}_{t}}{1000} e_{t} \right) \tag{Energy cost} \\
    &{} + 0.005 \left( \max_t \overline{P}_{t} \right)^2 \tag{Demand charge} \\
    &{} - \sum_{a_i} \left( d_i \cdot (\textit{value}_i - o_i\,\textit{penalty}_i) \right)  \tag{Once-off profit} \\
    \label{eq:objective} \tag{1}
\end{align}
Full mathematical formulation, including constraints, is provided in the reports
in Appendix A in the supplementary material, as presented by the participants.
More generally, constraints imposed by the problem include:
\begin{itemize}
    \item \textbf{Room availability:} Each activity must be scheduled in a room
    that is available and not double-booked.
    \item \textbf{Precedence constraints:} Some activities must occur before
    others, as specified by the precedence relationships.
    \item \textbf{Battery constraints:} The state of charge of the batteries
    must remain within their capacity limits at all times.
    \item \textbf{Activity timing:} Recurring activities must be scheduled
    within office hours (starting on or after 9:00 and finishing before 17:00).
    \item \textbf{Energy balance:} The total energy demand must be met by the
    combination of solar production, battery discharge, and grid supply,
    ensuring no feed-in to the grid.
\end{itemize}

Conceptually, the activity scheduling problem thus defined is an instance of the
Resource-Constrained Project Scheduling Problem~(RCPSP) with time
windows~\cite{Bartusch1988}. The RCPSP is a well-known problem in operations
research where the goal is to schedule a set of project activities within given
resource constraints and time windows. Here the activities are project tasks,
and the rooms and electricity use are the resource limits. Because recurring
activities with precedence must be scheduled on different days, the problem
also exhibits minimum time lags (the shortest allowable time between the start
of one activity and the start of another), as well as maximum lags (the longest
allowable time between the start of one activity and the start of another) due
to the limited time window of `daytime on weekdays' during which all recurring
activities must be scheduled. \citet{Bartusch1988} proved that even just testing
whether such a problem has a feasible solution is an NP-hard problem, meaning
that no \emph{efficient} solutions exist to solve it optimally (assuming the
widely held expectation that P $\neq$ NP holds, where P represents problems that
can be solved efficiently (in polynomial time), and NP represents problems for which solutions can be verified efficiently, in polynomial time).

Compared to the typical RCPSP with time windows, the problem also has a number
of additional considerations; access to an energy storage battery means that
some resource limits can be altered. And the once-off activities are
optional, while in RCPSP all activities must be scheduled. Furthermore, our
objective is not `shortest makespan schedule', but minimizing energy imports.
Minimizing energy imports refers to reducing the amount of energy that needs to
be purchased from the grid by optimizing the use of on-site renewable energy
sources and energy storage systems.

Despite the worst-case hardness of the RCPSP, it is known that randomly
generated instances may exhibit shallow hardness characteristics, meaning that
they are not overly difficult to solve and do not require extensive
computational resources. \citet{Vanhoucke2008} propose six topological
indicators of precedence graph connectedness (which measure the structural
properties of the graph, such as the number of nodes, edges, and the
connectivity between them), and perform a regression analysis on instance
hardness in terms of branch-and-bound search tree depth as a function of these
indicators. Insights from this work were used to construct an instance
generation algorithm tuned to generate instances that lie in the range between
over- and under-constrained, meaning that the problem formulation is neither too
easy nor too difficult to solve. This involved creating instances with
relatively few precedence constraints to ensure that the instance is close to parallel, and
having a mix of long chains of activities and free activities.

Feasible activity schedules were created as follows: 1) sample and schedule
activities, 2) assign precedence constraints, 3) set resource limits. In the
first stage, a number of activities are sampled, with duration $\mathcal{U}(2,
10)$ steps (from half-hour to two-and-half-hour long), number of rooms
$\mathcal{U}(1,3)$, using a small-sized room with
$\Pr(\text{small})=\frac{3}{4}$. Each activity is assigned power consumption
proportional to the maximum base power consumption (the highest power usage
recorded during the observation period) observed in the time series, sampled
from $\mathcal{U}(\frac{1}{20}, \frac{1}{10})$ of the maximum base load.
Once-off activities were given a value proportional to the average cost of
energy required, from $\mathcal{U}(0.9, 1.5)$ times the average cost. Sampled
activities were then assigned a day of the week, and an in-office-hours time of
day, constructing a tentatively valid schedule (meeting all the time-window
constraints), without any precedence constraints. In the second stage,
precedence constraints were sampled between scheduled activities such that they
were already satisfied by the tentative schedule: Each activity considers the
set of all activities on previous days, and samples without replacement from
this set a number of preceding activities chosen from a Binomial distribution
with $p=0.25$ (recurring) or $p=0.1$ (once-off). Thus, by construction there are
five bins of activities; those tentatively scheduled on Monday, having \emph{no}
precedence constraints, and those tentatively scheduled on Friday having many, with
potentially `long' arcs. Finally, in the third stage, the number of rooms was
determined as the maximum required by the tentative schedule of \emph{recurring
activities only}, meaning that once-offs have to fit in `gaps' different from
the tentative schedule by construction. 

Two sizes of instances were generated. Small instances had 50 recurring and 20
once-off activities, which is considered average in difficulty for the now
easily solvable~\texttt{psplib} set of benchmark instances~\cite{Kolisch1997}.
Large instances had 200 recurring and 100 once-off activities, nearly three
times the largest \texttt{psplib} instance and unlikely to be solvable to
optimality using `brute force' (i.e., trying all possible combinations). For
each of the two phases, 5 small and 5 large instances were generated, ensuring
that the competitors were required to solve the Phase~2 instances from scratch
(i.e., without opportunity to use warm starts (initial solutions based on
previous runs) or learned statistics about the Phase~1 instances).

\subsection{Evaluation of forecast accuracy and total energy cost}
\subsubsection{Evaluation of forecasts}
The forecasts of the 12 time series (energy demand of 6 buildings and power production from the 6 arrays of solar panels) were evaluated using the Mean Absolute Scaled Error (MASE) \cite{hyndman2006another}, a commonly used error measure for forecast
evaluation, which is defined as follows for a given series:
\begin{align*}
\mathit{MASE} &= \frac{\sum_{k=M+1}^{M+h} |{F_k - Y_k}|}{\frac{h}{M-S} \sum_{k=S+1}^{M} |{Y_k - Y_{k-s}}|}, \tag{2}
\end{align*}

\noindent where $M$ is the number of instances in the training series, $S$ is
the length of the seasonal cycle of the dataset, $h$ is the forecast horizon,
$F_k$ are the forecasts and $Y_k$ are the actual values. MASE was calculated
individually for each time series and averaged for the final error used to rank
submissions.\\

\subsubsection{Evaluation of optimal schedule and total energy cost}
Schedules were first checked for feasibility, after which the energy cost was
computed for feasible schedules.
\paragraph{Feasibility} Schedules were required to assign a time period to every
recurring activity while adhering to the following constraints, for every activity $a_i$: 
\begin{itemize}
    
    \item The starting period must be during the week having the first Monday of the month,

    \item Start time $\geq$ 9:00,
    \item Finish time (start time plus activity duration) $\leq$ 17:00,
   \item Activity precedence had to be observed.
\end{itemize}

\noindent Every battery schedule had to respect the capacity of the battery,
such that the State-of-Charge~(SoC) of the battery stays in $0 \leq
\textit{SoC}_t \leq \textit{capacity}$ for all time periods t.

\paragraph{Objective} For a feasible schedule, the objective value is computed
in terms of the cost of the schedule, which is to be minimized, using the
objective function~$O$ given in the previous section.

\subsection{Evaluation by the scientific committee and calculation of final scores}
\label{sec:sc_eval}

The 8 members of the scientific committee (SC) ranked the solutions using a form
inspired by peer review forms from machine learning conferences. The form
included free text criteria such as listing 3 advantages and 3 disadvantages of
the solution, commenting on the robustness of the optimization model, potential
generalizability of the approach, and potential overfitting in the approach. The
SC also ranked the solutions on a scale of (excellent/good/ok/poor) for each of:
scientific contribution, soundness, clarity, and reproducibility. Finally, the
jurors provided an overall evaluation of the submission on a scale of
(excellent/very good/good/acceptable/ok/poor). These scales were translated to
numerical values using a simple linear scale to produce a numerical score for
each participant, which was averaged over the SC members and ranked. The final
ranking of participants was calculated as the sum of $0.75$ of the energy
optimization ranking and $0.25$ of the SC ranking.

As well as submission of the 4-page report for the SC evaluation, participants
were required to submit their source code for verification by the organizers
that the code produced the reported solution. Participants were also required to
present their solution at a special session of the 2021 IEEE Symposium Series on
Computational Intelligence for further questions and checking by the panel and
audience. All shortlisted teams passed these hurdles without any problems.

\section{Solutions submitted, final rankings, and summary of results}
\label{sec:sol}

This section provides an overview of all submissions, the leaderboard timeline,
an overview of the best-performing solutions, and a more detailed evaluation of
the shortlisted solutions by the scientific committee.

\subsection{Submitted solutions}

In total, 49 individuals/teams participated in either Phase~1 or Phase~2 of the
competition. 36 individuals/teams submitted to Phase~1, and 36 (different)
individuals/teams submitted to Phase~2. 23 individuals/teams submitted to both
Phase~1 and Phase~2 of the competition. Many participants submitted multiple
times for evaluation. During Phase~1 there were 522 actual submissions
throughout the competition period. Approximately 50\% of teams attempted the
forecasting task only.

As it was not required to have submitted to Phase~1 in order to submit to
Phase~2, several new teams entered the competition for Phase~2 only. A number of
teams that were not competitive in Phase~1 dropped out of the competition before
Phase~2. Due to the challenging nature of the optimization problem, there were
fewer submissions (220) during Phase~2, since no feedback was given during the
competition period. 

Table~\ref{tab:topsolutions} shows the development of the leaderboard over time,
for Phase~1 and Phase~2. Table~\ref{tab:leadboard} shows the top positions of
the leaderboards at the conclusion of Phase~1 and Phase~2, respectively.

\begin{figure*}[htb]
	\centering

		\subfloat[Overall]{
		\includegraphics[width=0.32\textwidth]{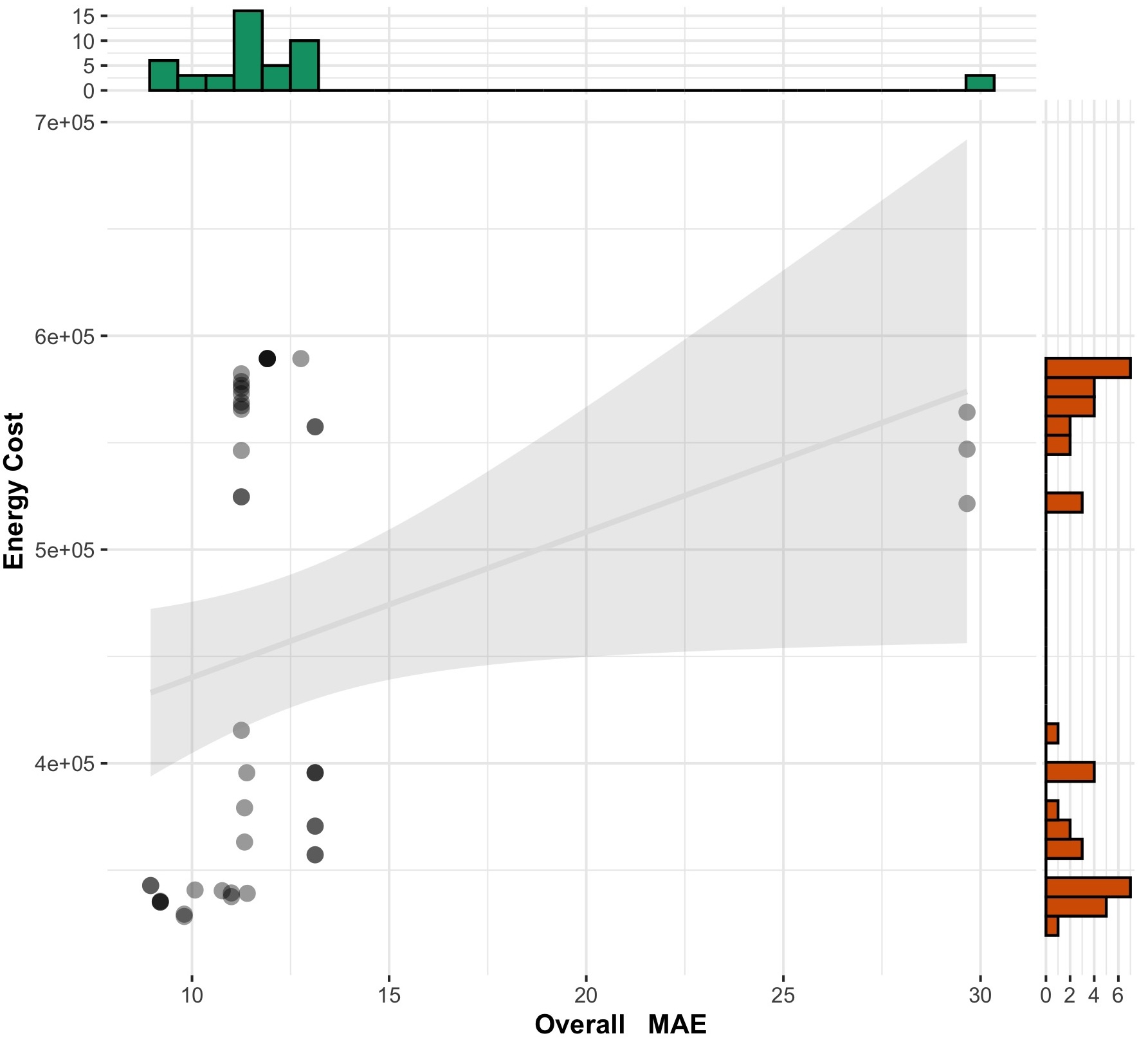}}
	\subfloat[Buildings]{
		\includegraphics[width=0.32\textwidth]{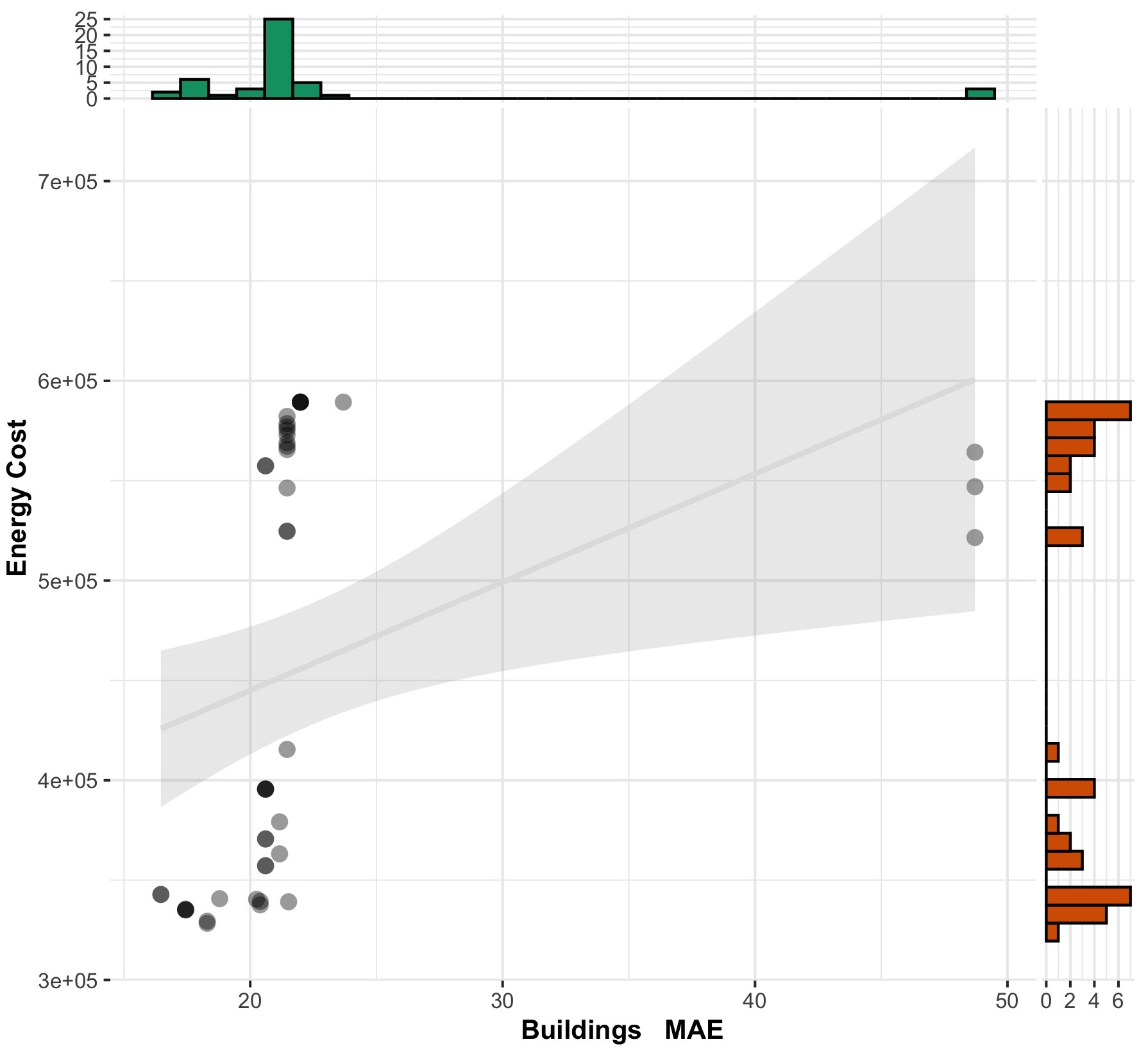}}
	\subfloat[Solar]{
		\includegraphics[width=0.32\textwidth]{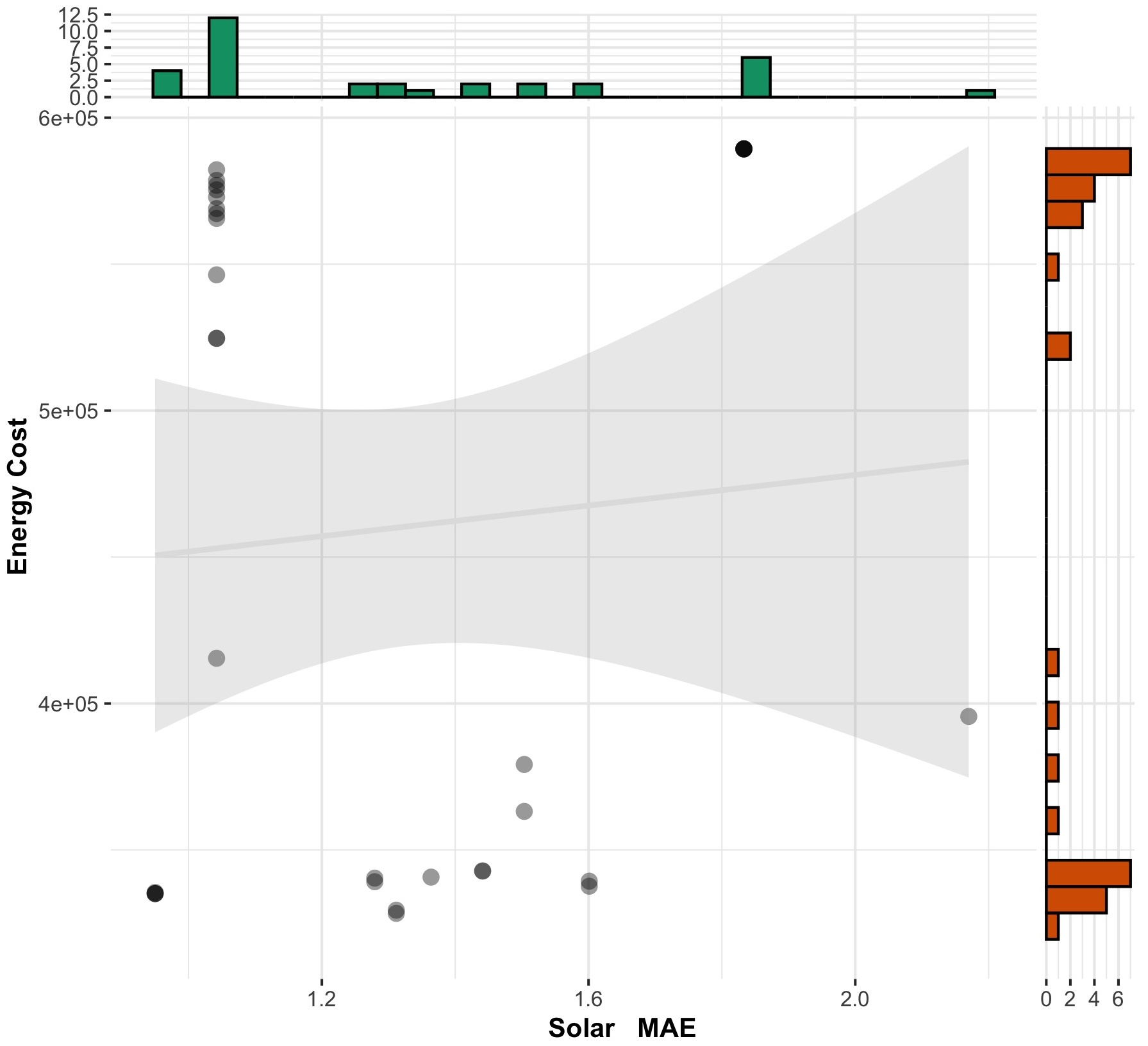}}

	\caption{Forecasting error (MASE and MAE) \textit{vs} energy cost for all solutions submitted to Phase~2 that outperformed the organizer-supplied baseline MASE and energy cost. The figure also shows Bayesian analysis results at the bottom. The natural logarithm of the Bayes factor $log_{e}(BF_{01})$ shows how strong the evidence is in favor of the null hypothesis over the alternative hypothesis. The posterior value $\hat{\rho}_{Pearson}^{posterier}$ and credible intervals $CI_{95\%}^{HDI}$ are estimated with $r_{beta}^{JZS}$ as the prior value.}
\label{fig:corr}
\end{figure*}

The relationship between forecasting and optimization performance, for solutions
in Phase~2 that outperformed the organizer-supplied baseline MASE and energy
cost, is shown in Figure~\ref{fig:corr}. The analysis has to be taken with
caution, as participants were not required to submit the forecast actually used
during optimization, meaning that the actual forecast reported may not have been
that used. Furthermore, the linear fits shown as lines in the plot do not
represent the data well and serve only as an overall guidance. However, it is
immediately apparent that there is very little correlation between solar
forecast accuracy and energy cost. This is because solar power generation is
approximately an order of magnitude smaller than the actual building load
meaning that solar energy forecast errors only have a small effect on total
energy costs. Forecasting building energy demand was much more important to
total cost, hence the higher correlation between the two compared to solar. The
correlation is particularly distinct for the Mean Absolute Error (MAE), as
shown in Figure~\ref{fig:corr}.

\begin{table}[htbp]
  \centering
  \caption{Best solutions over time during Phase~1 and Phase~2.}
    \begin{tabular}{lrr}
    \toprule
    \textbf{Date} & \textbf{Best Cost} & \textbf{Best MASE}\\
    \midrule
Phase~1\\    
    \midrule
    19/07/2021 & 453,317 & 1.1365 \\
    16/08/2021 & 453,317 & 0.8776 \\
    30/08/2021 & 445,218 & 0.8776 \\
    13/09/2021 & 444,858 & 0.8106 \\
    27/09/2021 & 444,858 & 0.6625 \\
    11/10/2021 & 439,071 & 0.6320 \\
    \midrule
Phase~2\\    
    \midrule
    14/10/2021 & 339,160 & 0.8030 \\
    18/10/2021 & 339,160 & 0.8030 \\
    23/10/2021 & 337,625 & 0.6927 \\
    27/10/2021 & 337,625 & 0.6927 \\
    30/10/2021 & 329,441 & 0.6927 \\
    02/11/2021 & 328,359 & 0.6460 \\
    \bottomrule
    \end{tabular}
  \label{tab:topsolutions}
\end{table}

\begin{table}[htbp]
  \centering
  \caption{Top of the leaderboard after Phase~1 and Phase~2.}
    \begin{tabular}{lrr}
    \toprule
    \textbf{Team} & \textbf{MASE} & \textbf{Energy cost (\$)}\\
    \midrule
Phase~1\\    
    \midrule
    MA\&RE & 0.982255 & 439,071 \\
    HRI & 0.658880 & 439,936\\
    RB & 0.632086 & 446,416\\
    
    FRESNO & 0.777158 & 482,870 \\
    AS & 0.695587 & 483,643\\
    QSZU-PolyU & -- & 485,733\\
    
    EVERGi & -- & 710,227 \\
    \midrule
Phase~2\\    
    \midrule
    MA\&RE & 0.744052 & 328,359 \\
    RB & 0.646022 & 335,107\\
    HRI & 0.855737 & 339,160\\
    EVERGi & 0.807299 & 340,726 \\
    QSZU-PolyU & 0.774996 & 342,810 \\

    FRESNO & 1.870326 & 357,210 \\
    AS & 0.847391 & 363,168 \\

    \bottomrule
    \end{tabular}
  \label{tab:leadboard}
\end{table}

\subsection{Overview of best-performing solutions}
\label{sec:bestsol}

\begin{table*}[htbp]
  \centering
  \caption{Summary of optimization methodologies of shortlisted solutions.}
    \begin{tabular}{lr|lll}
    \toprule
        \hline
    \multicolumn{5}{c}{\textbf{Optimization methodology}}\\
    \hline
    \textbf{Team} & \textbf{EC (\$)} & \textbf{Algorithm} & \textbf{Software} & \textbf{Comments}\\    
    \hline
    \hline
    MA\&RE & \textbf{328,359} & MIP/MIQP & Gurobi & Sample Average Approximation Method (SAAM) is employed \\
   &&& &  in which the optimization model minimizes the average cost\\
      &&& &  of a solution over multiple scenarios\\
    \hline
    RB & 335,107 & MIP/MIQP & Gurobi & Two-staged process\\
        \hline
    HRI & 339,160 & MIP/MIQP & Gurobi & Split into three sub-problems, use linearization technique\\
    \hline
    EVERGi & 340,726 & CMAES or GA and & Gurobi for MIP, & Evolutionary algorithms for activity scheduling, \\
    && subsequent LS for schedule, & pygmo for CMA-ES, &  MIP for battery scheduling \\
    && MIP for batteries & PyGAD for GA & \\
    \hline
    FRESNO & 357,210 & MIP & Gurobi & Linearization, did not schedule once-off activities \\
        \hline
    QSZU-PolyU & 342,810 & LS (Local Search) & -- & Develop a custom method to generate feasible solutions, \\
    &&&& randomly modify those \\
        \hline
    AS & 363,168 & MIP & Gurobi & Large Neighborhood Search coupled with scenario-based \\
     &  &  &  &  robust optimization, fix-and-optimize approach \\    
    \hline
     \bottomrule
    \end{tabular}
  \label{tab:opt_summary}
\end{table*}
    
\begin{table*}[htbp]
  \centering
  \caption{Summary of forecasting methodologies of shortlisted solutions. Errors reported are averages over all building demand series and over all solar production series, respectively.}
\footnotesize
    \begin{tabular}{lllr|lll}
\toprule    
    \hline
    \multicolumn{7}{c}{\textbf{Building demand forecasting methodology}}\\
    \hline
        \textbf{Team} & \textbf{MASE} & \textbf{MAE} & \textbf{RMSE} & \textbf{Algorithm/Software} & \textbf{Input features} & \textbf{Comments}\\    
    \hline
    \hline
    MA\&RE & 0.841 &18.294 &27.265   & Ensemble of LightGBM & Calendar features, daily/hourly & Ensemble over models that use  \\
    && &&& weather data, rolling statistics &  daily, weekly, and daily\&weekly  \\
    && &&&  &  weather features  \\
        \hline
    RB & \textbf{0.807}  &17.441 & 25.263  & Quantile regression forest & Calendar features, Fourier terms, & Groups of buildings trained to- \\
    && & & from R package ``ranger'' & BOM data, ERA5 data, lagging & gether as they were observed to  \\    
   && & & & and leading features & be closely correlated over time.\\    

        \hline
    HRI & 1.089 &21.522 & 31.029  & Seasonal median forecast & No external inputs & Eight weeks of historical data  \\
   && & & over last 8 weeks & & as input \\    
        \hline
    EVERGi & 0.959 &18.790 & 26.594  & LightGBM & Calendar features, weather data, & Log transform as preprocessing, \\
    && &  &  & occupancy rates, lags, seasonality &  Prophet for feature engineering; \\    
    && &  &  & and trend features &  Building 4 is treated as \\
    && &  &  &  &  a multi-class classification\\
        \hline
    FRESNO & 0.921  &20.608 & 28.840
 & STL decomposition, then  & Calendar features, occupancy, & 2 months of historical data  \\
   && & & ARIMA, RF, LightGBM, & hourly weather data & as input \\    
   && & & and SVM & & \\     
        \hline
    QSZU- & 0.835  &\textbf{16.460} & \textbf{22.751}  & Different ML models,  & Calendar features (hour, minute, & Models trained across buildings, \\
   PolyU & & & & including neural networks & weekday), total energy demand &  preprocessing different for each  \\    
   && & & & of Victoria & building \\    
        \hline
    AS & 0.945  &21.164 & 29.965  & Random Forest, Quantile & Weather data, calendar effects, & -- \\
     &&   & & Regression Forest & impact of COVID-19 restrictions, & \\
     &&   & & & exams period, and others & \\

    \hline
    \hline
    
    \multicolumn{7}{c}{\textbf{Solar production forecasting methodology}}\\
    \hline
        \textbf{Team} & \textbf{MASE}& \textbf{MAE} & \textbf{RMSE} & \textbf{Algorithm/Software} & \textbf{Input features} & \textbf{Comments}\\    
    \hline
        \hline 
    MA\&RE & 0.647  &1.312 & 2.309   & Ensemble of LightGBM & Calendar features, daily weather & Ensemble over models that use \\
 &&  & & & data, hourly weather data, & daily, weekly, and daily\&weekly  \\    
 &&   & & & various rolling statistics & weather features \\    

        \hline
    RB & \textbf{0.485}  & \textbf{0.950} & \textbf{1.855}   & Quantile regression forest & Weather data, leading and & All of the solar instances  \\
   &&    & & & lagging features & were trained together \\    
        \hline
    HRI & 0.623  &1.279 & 2.397   & Random forest & Weather data, leading and & --\\
    &&     & &  from scikit-learn & lagging features & \\   
        \hline
    EVERGi & 0.656  &1.364 & 2.412   & LightGBM & Calendar features, weather data, & -- \\
  &&     && & mean value at similar time & -- \\    
        \hline

    FRESNO & 2.820  &5.639 & 9.249  & ResNet, Refined Motif (RM) & Solar generation data, weather, & Team submission was erroneous, \\    
  &&     && & date time & error would be lower\\

        \hline
    QSZU- & 0.715   &1.441 & 2.555    & Ensemble of various different & Surface solar radiation most & -- \\
 PolyU & & & & types of neural networks,  &   important feature & \\
 &&   & & SVR, Prophet & & \\
    \hline
    AS & 0.750 &1.504 & 2.617    & Ensemble of Random Forest, & Weather data, calendar features & -- \\
 &&   & & Gradient Boosting Machines, & & \\
 &&   & & Ridge Regression, and Local & & \\
 &&   & & Learning Regression & & \\
    
        \hline
    \bottomrule
    
    \end{tabular}

  \label{tab:fc_summary}
\end{table*}

Tables~\ref{tab:opt_summary} and \ref{tab:fc_summary} present an overview over
the optimization and forecasting methods used by the shortlisted teams. It is
evident in Table~\ref{tab:opt_summary} that most teams used mixed integer
programming (MIP), or mixed integer quadratic programming (MIQP) with linear
relaxations for the optimization. Only the teams ranked at 1st and 7th place
considered forecast uncertainty by predicting scenarios and employing stochastic
or robust optimization. The other teams relied on point forecasts and
deterministic optimization. Regarding software, most teams used Gurobi for
optimization via a Python interface. Some notable exceptions were the EVERGi
team, who used evolutionary algorithms, in particular CMAES, combined with a
subsequent local search, for the activities schedule. Team QSZU-PolyU used a
simple heuristic approach for the scheduling that was then further optimized
with a local search, which provides an excellent benchmark from which to assess
possible gains obtained by more complex methodologies.

Table~\ref{tab:fc_summary} shows that most of the top performing teams used
tree-based algorithms, namely LightGBM and Random Forest (RF), to forecast
building energy demand. A notable exception was the team from the Honda Research
Institute (HRI, consisting of Steffen Limmer and Nils Einecke), who were able to
achieve good results with a very simple technique: a seasonal median of demand
over the previous 8 weeks. Several teams observed that Building 4 had very low
demand, which led the EVERGi team to model this building as a multi-class
classification problem. Other teams employed simple techniques, e.g., Richard
Bean (RB) used a median forecast for this building. There are considerable
amounts of missing values in the data, which likely contributed to the success
of tree-based methods. Most teams used the weather data provided (daily and
hourly), together with calendar features and/or Fourier terms. Weather data was
used with both lagging and leading features since the competition assumed the
availability of a perfect weather forecast. Some teams used other features such
as the total energy demand for the state of Victoria, and occupation rates as
estimated from COVID-19 restriction information and the academic calendar (e.g.,
exam periods).

Tree-based algorithms such as LightGBM and Random Forests were employed by most of the top solutions for solar forecasting. In particular, the two best solutions in terms of forecasting accuracy are based on these. Other approaches were neural networks such as ResNet (FRESNO team), and ensembles that included support vector regression (SVR), Prophet, Ridge Regression, and other algorithms. The features used by the participants were again lagging and leading weather features (solar irradiation in particular), and calendar features.

\subsection{Evaluation of results by the scientific committee}
\label{sec:sc_results}

\begin{figure}
\centering
	\begin{center}
 		\includegraphics[width=0.5\textwidth]{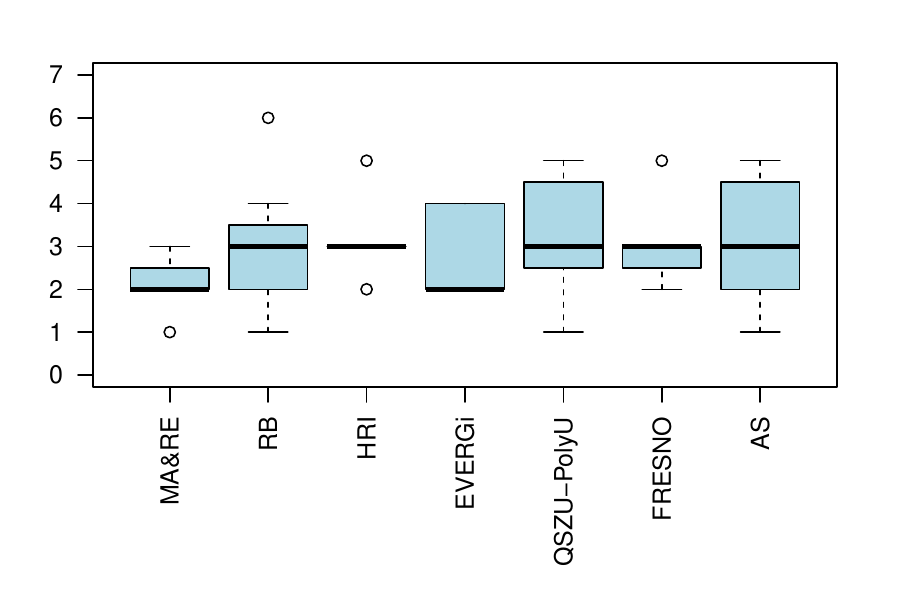}
 	\caption{\label{fig:scoverall} Overall evaluations by the scientific committee. Lower values are better, as the y-axis is the rank.}
\end{center}
\end{figure}

Figure~\ref{fig:scoverall} shows the overall evaluation of the shortlisted teams
by the scientific committee (SC). The average score in each subcategory for
these teams is shown in Table~\ref{tab:SC_ranks}. Results show that, generally
speaking, the highest-ranked submissions were those that gave the best solutions
in terms of energy cost. However, when the SC evaluations were incorporated with
energy cost rankings, 5th and 6th ranked teams swapped, and the 3rd and 4th
ranked teams were ranked equally at 3rd place. A summary of the evaluations is
given in the beginning of each team's detailed description in Section IV. Further details of
the evaluation of each team by the scientific committee are given in the
appendix of the paper, in the supplementary material.

\begin{table}[htb]
\centering
\begin{tabular}{rrrrrrrr}
  \toprule
 & MA\& & RB & HRI & EVE- & QSZU- & FRE- & AS \\ 
  & RE &  & & RGi & PolyU & SNO &  \\ 
  \midrule
  Sc.\@ Contrib.\@ & 2.12 & 2.12 & 2.62 & 2.50 & 2.14 & 2.14 & 2.29 \\ 
  Soundness & 1.50 & 1.75 & 2.00 & 1.75 & 2.14 & 2.00 & 2.00 \\ 
  Clarity & 1.62 & 2.25 & 2.00 & 1.88 & 2.43 & 1.86 & 1.71 \\ 
  Reprod. & 2.12 & 2.50 & 2.12 & 2.12 & 2.29 & 2.29 & 2.29 \\ 
  \midrule
Overall & 2.12 & 3.00 & 3.12 & 2.75 & 3.29 & 3.00 & 3.14 \\ 
   \bottomrule
\end{tabular}
\caption{Average evaluations by the scientific committee on each criterion. Ranking was from 1: excellent, to 4: poor, for the first 4 items, and from 1: excellent, to 6: poor for the overall evaluation.}
\label{tab:SC_ranks}
\end{table}

\section{Descriptions of best-performing solutions}
\label{sec:det_desc}
Summaries of the 7 shortlisted solutions are presented below, in the order of
their final score, i.e., the winning solution is presented first, etc. For more
details, refer to the appendix of the paper, in the supplementary material.

\subsection{Mahdi Abolghasemi and Rasul Esmaeilbeigi's solution (MA\&RE)}
\textbf{Summary of the Scientific Committee:}
An ensemble of LightGBM models with calendar features and dynamic features
(lags, mean, standard deviations), with ensembling over models that use daily,
weekly, and daily and weekly weather features. Optimization over multiple
scenarios, with sample average approximation that minimizes average cost over
multiple scenarios. The approach seems reproducible and solid, although classic.
It is a fast and accurate forecasting and optimization.

\noindent \textbf{Advantages:}
Easy, systematic, robust, reproducible methodology. A rigorous problem
formulation so that the model is not unduly complex. A good exploration of
alternative forecasting techniques. Furthermore, a stochastic optimization
approach that uses multiple forecasts. Also, a large neighborhood search and
good decomposition technique.

\noindent \textbf{Disadvantages:}
Somewhat ad-hoc hyperparameter tuning, a focus on ``local'' models for
forecasting that work on every series separately. The algorithm choice is not
clearly motivated, and some manual and unclear steps are in the forecasting
methodology.

\noindent \textbf{Robustness:}
Though some of the forecasting seems ad-hoc, e.g., the choice of size of
training set per series, and optimization is typically not directly
transferrable, the general methodology is highly generalizable and robust. It
can be applied to similar problems with minor adjustments. Overfitting is
adequately addressed with cross-validation, L1 regularization and early
stopping.
\newline
\newline
This solution ranked 1\textsuperscript{st} in the optimization and
2\textsuperscript{nd} in the forecasting challenge of the competition. An
extensive exploratory data analysis was conducted to look at trends,
seasonality, and intermittency patterns of the data. The impacts of COVID-19
were explored on buildings' power demand since part of the provided data and the
forecasting horizon was during the pandemic. Since both solar power and
buildings' demand are highly dependent on weather conditions, the hourly weather
data provided by the organizers of the competition and downloaded daily data
from the BOM website~\cite{bom-climate} were used. Various statistical and
machine learning models including seasonal ARIMA, RF, LightGBM, and SVR were
explored for building the predictive models. While the generated forecasts
especially with RF and SVR were fairly accurate and competitive to LightGBM,
LightGBM was opted for since it is significantly faster and returns reliable
forecasts. The forecasting models were all trained with LightGBM where calendar
features, daily weather data, hourly weather data and various rolling statistics
of these features were used as input variables in the model. Hyperparameters
were optimized and the most significant features for each model were selected.
Several forecasts were generated with different models to provide a larger pool
of scenarios for the optimization part of the competition. The final submitted
forecasts were an ensemble of two LightGBM models for each series where daily
and hourly features were used for each series and hyperparameters were
optimized. 

The objective function of the optimization part minimizes the total energy cost
that includes the square of the maximum load, i.e., a quadratic term. A
linearization technique was used to linearize this objective function and
develop a mixed integer linear program that captures all constraints of the
problem. One of the input parameters of the optimization model is the
\textit{net base load}, which is the difference between the total predicted base
load of the buildings and the generation of their solar panels, per time slot.
The proposed optimization approach does not rely on one forecast.
Decision-making under uncertainty plays a crucial role in managing energy
systems. Uncertainty should be addressed properly since these systems are highly
reliant on the predetermined energy prices and policies as well as predictable
energy loads and demands \cite{dai2021utilization}. In this solution, the net
base loads of each time slot are considered as a random variable in the
optimization model and the so-called Sample Average Approximation Method (SAAM)
is employed in which the optimization model minimizes the average cost of a
solution over multiple scenarios (predictive outcomes) rather than just one. The
final submission employs 6 forecasting scenarios. This approach generally
prescribes a solution with least expected cost that is also less sensitive to
the forecasting errors. See \citet{esmaeilbeigi2021benders} for more details of
SAAM. The code is publicly available from GitHub for
forecasting\footnote{\url{https://github.com/mahdiabolghasemi/IEEE-predict_optimise_technical_challenge}}
and
optimization.\footnote{\url{https://github.com/resmaeilbeigi/IEEE_CIS_3rd_Technical_Challenge_Optimiser}}
For more details of this solution see Appendix A  in the supplementary material.

\subsection{Richard Bean's solution (RB)}
\textbf{Summary of the Scientific Committee:} The submission uses a quantile regression forest with weather
data from BOM and ERA5 for forecasting. Models are built on groups of series
(buildings) rather than per series. Various different approaches are used for
optimization. Both MIP and MIQP are used in a two-stage approach to tackle the
quadratic objective. The forecasting uses a good selection of covariates, with
justifications. The general methodology is very specific with great
understanding of the data, but also sometimes seems ad-hoc, to the point of
forecasting manually chosen constants in some instances. It achieves very
accurate forecasting results, which are the best in the competition.

\textbf{Advantages:} Highly effective in forecasting, achieving high accuracy.
Performs global modelling across series. Optimizes the start date for the
training data. Deals with outliers (although manually), and has a good rationale
for variables used, with careful analysis of the data. The components seem well
integrated.

\textbf{Disadvantages:} Some steps are ad-hoc, and it is not clear why some of
them are carried out. Some decisions are not well justified. This may lead to a
lack of reproducibility. For example, setting manual thresholds for outlier filtering,
grouping of buildings based on observation. Parameter optimization was performed
fully against Phase~1, no time series cross-validation was performed. Some parts
of the optimization approach are very heuristic.

\textbf{Robustness:} As some steps seem ad-hoc, extensions to other scenarios
may require adaptations. Most design choices have been made based on the
particular data. As such, it is a quite specific solution. Though the particular
approach will not be easy to generalize, the main idea can be generalized.
\newline
\newline
Energy consumption by buildings, and solar energy production was forecast using random forests. Inputs to these models consisted of historical building use and weather data provided by the BOM and the European Centre for Medium-Range Weather Forecasting (ECMWF). Forecast accuracy was improved by thresholding energy consumption for buildings when these appeared as outliers. The forecast window was varied to determine the window length that maximised forecast accuracy. Manual feature selection was used to identify the most important features for each type of forecast and reduce the likelihood of over-fitting. Optimization was performed using a MIP solver to minimize peak load due to recurring activities. One-off activities were then incorporated into the schedule and a MIQP was used to shift activities to minimize total cost, using a \emph{no forced discharge} battery management policy.
The source code of this solution is available online.\footnote{\url{https://github.com/RichardBean/IEEE-Predict-Optimize-Challenge}} 
For more details of this solution see \citet{Bean2022Forecasting}, and for the evaluation of the approach by the scientific committee see Appendix B in the supplementary material.

\subsection{HRI Team's solution}
\noindent \textbf{Summary of the Scientific Committee:} The solution is based on simple median
values for the load prediction, Random Forest for solar power production, and
(an efficient) MILP for the optimization task. It therewith assumes that the
influence of the weather forecast on the load is only marginal. It performs an
elegant decomposition of the optimization problem with a variety of different
settings to determine the final ``best'' setting. It is overall a simple,
straightforward application of existing technologies that achieves results close
to the best-performing methods both in forecasting and optimization.

\noindent \textbf{Advantages:} The approach is designed to achieve a solution in
an acceptable time frame. Random Forest is robust for prediction. The
participants experimented with different levels of decomposition for the
optimization, with a good problem formulation, namely linearization by using
peak load instead of the quadratic function to schedule recurring activities,
and by separating the optimization into the same two steps as the competition
winners, namely assigning buildings to activities as a second step.

\noindent \textbf{Disadvantages:} There have been some manual decisions on what
data to use, and the load prediction seems a bit too simplistic, although it
achieves a decent performance. No comparison with other forecasting methods was
given, and no discussion of why linearization of the objective does not lead to
a decrease in solution quality. Overall, the results are good but not excellent.

\noindent \textbf{Robustness:} The methodology seems robust and minor adaptation
would be required for other scenarios, although outlier filtering is done
manually. The simplicity of methods and use of standard procedures makes the
method relatively easily generalizable.
\newline
\newline
For the load prediction, a simple statistical approach was used, which predicts the load at a certain time step of a week as the median over the load values at the corresponding time steps of eight weeks of historical data.
The PV production was predicted with a machine learning approach based on an RF with 14 input features (mainly weather data). For the optimization, a combination of mixed integer linear programming and mixed integer quadratic programming together with the Gurobi solver was employed. In order to accelerate the optimization, different measures were applied:
\begin{enumerate}  
 \item The activities were assigned to buildings in a step performed separately from the main optimization.
 \item The number of decision variables was reduced by excluding start times of activities, which are infeasible with respect to precedence constraints. 
 \item The setting of the parameters of the employed solver was tuned.
 \item The problem was decomposed into three easier subproblems.
 \item The objective function was linearized.
\end{enumerate}
A more detailed description of the approach as well as a thorough evaluation of it on the data and problems of the second phase of the challenge can be found in \citet{limmer2022efficient}. 
For the evaluation of the approach by the scientific committee see Appendix C  in the supplementary material.

\subsection{EVERGi Team's solution}
\noindent \textbf{Summary of the Scientific Committee:} The
approach uses an evolutionary algorithm for initial scheduling of activities,
followed by a local search, and MIP for the batteries. Seasonal and trend
decomposition (STL, Prophet) followed by LightGBM is used for forecasting.

\noindent \textbf{Advantages:} The methodology uses good
preprocessing that, for example, found drift in the demand of Building~5. It
uses a good forecasting methodology that applies transformations and
decompositions. The optimization is a combination of heuristic and complete
solvers, and as such a novel optimization idea that seems to work well.

\noindent \textbf{Disadvantages:} Different buildings are
treated differently in the forecasting. The computation time will presumably be
high and the activity schedule might be slow. The evolutionary algorithms seem
somewhat ad-hoc, and the schedule improvement with Gurobi adds complexity to the
model.

\noindent \textbf{Robustness:} The method is a relatively general and robust
approach that seems applicable in other settings with some minor adjustments.
\newline
\newline
The optimization methodology consisted of the following elements:
Multi-dimensional time series forecasting using LightGBM \cite{ke2017lightgbm}
was performed to predict both future energy production and consumption from
historical data. Features used for modelling included seasonality and trend,
weather data, calendar features, time of day, academic calendar, and
proportional occupancy of buildings. A log transform was used to reduce
variability in consumption. The optimal schedule of room usage and battery
storage was created by first creating a base schedule of recurrent and one-off
activities that satisfied precedence and room availability constraints. This was
performed using both a Genetic Algorithm and the Covariance Matrix Adaptation
Evolution Strategy (CMA-ES) \cite{hansen2003reducing} independently. The base
schedule was then improved by modifying the times of activities one-by-one when
this would reduce costs. Recurrent activities were evaluated first, followed by
one-off activities, in order of precedence for each group. The optimal battery
schedule was determined using MIP implemented in the Gurobi solver. The final
submission had an error so that only one battery was used instead of two. After
fixing this error, the method would have obtained the second-lowest cost in the
competition. The source code of this solution is available
online.\footnote{\url{https://github.com/ujohn33/EVERGI_predict_optimize}} A
more detailed description of the approach can be found in
\citet{ruddick2022evolutionary}. For the evaluation of the approach by the
scientific committee see Appendix D in the supplementary material.

\subsection{FRESNO Team's solution}
\noindent \textbf{Summary of the Scientific Committee:} The paper uses STL
decomposition, followed by a separate forecast of each time series with ARIMA,
Random Forest, Gradient Boosting, and SVM. For the solar panels, ResNet is
trained using Refined Motif, proposed by the participants in another paper. The
optimization is then done using MIP with a sensible decomposition, that solves a
Linear Program relaxation first, then the scheduling problem.

\noindent \textbf{Advantages:} The approach uses a systematic
forecasting methodology. It uses a good Linear Programming relaxation to bound
the optimization problem, that focuses on the peak demand. This appears to make
the optimization more robust w.r.t. the forecast quality, as the forecasts are
not very accurate.

\noindent \textbf{Disadvantages:} Some buildings are not STL
decomposed and treated differently, with no justifications. ResNet might need
more data than available here. In general, the forecasting is not as accurate as
the methodologies of other participants.

\noindent \textbf{Robustness:} Besides the complexity of the procedure and the
many methods involved, the methodology seems to be able to be used on other
datasets.
\newline
\newline
As the building and the solar patterns were completely different, two different
sets of forecasting models were developed. Various research on forecasting
techniques shows that ensemble methods outperform individual ones in many
cases. Therefore, the voting regressor from the Python sklearn package was used
to forecast the buildings' load. This regression model fits several estimators
on the same dataset and then averages them out to get the actual predictions. It
was found that tree-based methods like RF and gradient boosted trees gave the
highest accuracy for this dataset. Also, to capture the cyclic and seasonal
variations of the buildings' load, STL decomposition was incorporated with the
above methods to improve prediction accuracy.

Solar generation by its seasonal nature, tends to have repeated patterns. As a
result, it might be useful to extract the most repeating pattern from the solar
time series data and account for variances from the baseline using exogenous
variables such as weather data. This repeating pattern is discovered by a
refined motif (RM) method which is developed by the competition participants in
The discovered repeating patterns along with other exogenous variables were fed
to a 1D convolutional neural network (1D-CNN) during Phase~1 to make
predictions. Over-parameterization of CNNs can yield better performance, but
training is costly in terms of computation time \cite{Zerveas2020}. Thus,
Residual Networks (ResNet) were implemented as an option for an NN that is deep
but also has comparably low computational cost. The performance of the ResNet
model was generally better compared with 1D-CNN. Note that the submission of the
solar forecasts for Phase~2 was erroneous. A corrected calculation for Phase~2
should give comparable MASE values to \citet{yuan2021irmac}.
Phase~1.\footnote{\url{https://gitlab.com/ryuan/ieee-cis-data-challenge-fresno/-/blob/main/Solar_prediction.ipynb}}

For the optimization part of the competition, to capture the constraints of the
scheduling problem, binary variables are necessary, for example at which
interval a particular task is active. Thus, MIP was used to model this problem.
From the problem description the following challenges were identified:
\begin{enumerate}
    \item Scheduling for one month with 15-minute granularity means vectors of
    size 2880. Hence, using more activities leads to exponentially increasing
    complexity.
    \item The peak power cost involved a square term making this problem a MIQP.
    \item For best economic benefit, it was necessary to schedule all activities
within working hours. This also contributes to the peak power term, which is a
sizeable chunk of the energy cost. \end{enumerate} \par These challenges heavily
influenced the tractability of the problem. It was also found that the maximum value
obtained by scheduling non-recurring activities was an energy cost of
approximately \$16,000 and this may not be worth the extra cost and computation
required to schedule these tasks. Accounting for this, the following two steps
were used to simplify the problem: 
\begin{enumerate}
    \item Only recurring activities were modelled in the problem. 
    \item The problem was converted to a mixed integer linear program by setting
    a limit on the peak power term over the month and removing it from the
    objective.
\end{enumerate}

The methodology was hence divided into 4 sub-sections: data pre-processing,
building load forecasts, solar generation forecasts, and optimal scheduling
problem. The code of this solution is available
online.\footnote{\url{https://gitlab.com/ryuan/ieee-cis-data-challenge-fresno}}
For more details see Appendix E  in the supplementary material and
\citet{Kumar2022Optimal}.

\subsection{QSZU-PolyU-Team's solution}
\noindent \textbf{Summary of the Scientific Committee:} 
Forecasting is performed using a variety of ML models (including neural
networks). Single models are trained for all buildings, but preprocessing is
different for each building. Then, this team uses a bi-level optimization to
first identify an optimal timetable using local search, with relaxation. After
this battery scheduling is optimized, again with a local search method, based on
the optimal timetable. This is an ad-hoc and very simple optimization algorithm,
which is a useful approach to benchmark the effectiveness of sophisticated MIP
methods against this simple alternative.

\noindent \textbf{Advantages:} 
A single forecasting model is built for all buildings, using weather data. The
optimization is very simple. The observation that forecasting accuracy may not
have a large impact on quality of optimization is interesting and useful.

\noindent \textbf{Disadvantages:}
The method presumably needs expensive training, and no hyperparameter tuning has
been discussed. Only 2 months (August and September) are used for training.
There is different ad-hoc preprocessing for different buildings. The forecasting
accuracy is weak overall. Local search does not allow to judge solution quality.

\noindent \textbf{Robustness:} 
The preprocessing seems not generalizable, but the rest of the methodology can
be adapted to other scenarios with minor changes.
\newline
\newline
Optimization was performed by first determining an optimal timetable after which
the battery use was scheduled. Since activities to be timetabled had precedence
relationships, a feasible set of activities that could be performed each day was
constructed. Local search was then applied to this feasible set to determine the
optimal schedule. Batteries were assumed to be in one of three states: hold,
discharge or charge. The optimal battery state at each time slot was determined
again using local search. Weather forecasts were made at 15-minute intervals
from historical data. Total energy use across all buildings was found to
correlate with total Victorian energy use. The prediction of energy use within
individual use varied between buildings. Consumption data for some buildings
contained many missing values, so consumption was set at constant levels. For
other buildings' consumption was predicted by time of day, and by week day or
weekend. Energy production was forecast from surface solar radiation data.

The source code of this solution is available online.\footnote{\url{https://github.com/xuyaojian123/IEEE-Predict-Optimize-Challenge}}
A more detailed description of the approach can be found in \citet{Zhu2021local}.

\subsection{Akylas Stratigakos' solution (AS)}
\noindent \textbf{Summary of the Scientific Committee:} This participant used ML
forecasting methods such as Random Forest, Gradient Boosting Machines,
regression variants to predict PV power generation. Building demand forecasts
were created using Random Forest models and quantile regression, using calendar
and weather features. The optimization problem was solved using MIP via Gurobi.
The novelty in the proposed method is the use of a fix-and-optimize approach,
whereby sections of a feasible search space are ``fixed'' while the solver
explores the remaining free variables.

\noindent \textbf{Advantages:} The method uses well established ML models for
forecasting both power production and building demand. The ``fix and optimize''
nature of the solver solution has the potential to increase performance speed.
Combining these elements creates an effective solution tool with a
straightforward data flow/solution path. Thus, the optimization approach is
robust and easy to generalize. Minimization of the worst case expected cost
helps hedge against large forecast errors.

\noindent \textbf{Disadvantages:} The forecasting seems to not have received as
much attention as in other solutions, and this may have had some influence on
the results. The search in optimization is a bit greedy, and there will be
degradation of the solver solution using ``fix and optimize'' compared to a more
exhaustive solution.

\noindent \textbf{Robustness:} The scenarios are based only on Building~3, which
makes the approach less generalizable. Apart from this, very few assumptions are
made about the input data etc., so that this approach seems highly applicable to
other settings, and the solution delivers many insights that can help to adapt
it.
\newline
\newline
The proposed solution was guided by several challenges that revealed themselves
during the early stages of the competition. First, the limited computational
resources did not allow to solve the (multiple) problem instances to optimality.
Second, the computational cost also hindered our ability to explore different
strategies during the validation phase, e.g.\@ how to tackle the parameter
uncertainty. Lastly, as the time to be allocated in this challenge was also
limited, the decision was made to focus on the optimization component at the
expense of the prediction component. Considering the above, the proposed
solution adheres to the following: \emph{(i)} can be implemented in a standard
machine, \emph{(ii)} provides competitive results relatively fast, and
\emph{(iii)} provides hedging against large forecast errors.

To this end, the solution was based on a fix-and-optimize heuristic search to
iteratively improve an initial solution of the MIP solver
(\textit{matheuristic}). The problem was formulated as a large MIP and the
proposed solution combines Large Neighborhood Search coupled with scenario-based
robust optimization for handling uncertainty in the objective function. The
uncertainty in problem parameters (i.e., renewable production and electricity
demand) was modeled with scenarios based on marginal predictive intervals. A
robust objective was then formulated to minimize the worst-case cost within the
set of scenarios, thus offering protection against miscalibrated forecasting
models. The solution methodology then considered the following steps. First, an
adequate feasible schedule was derived considering only hard problem
constraints, in this case the scheduling of recurring lecture activities. Next,
the solution was improved iteratively with a fix-and-optimize heuristic search.
At each iteration, the MIP solver explored a large neighborhood by fixing a
subset of variables and optimizing over the remaining free variables. The
process was repeated several times until a stopping criterion was met. Code for
this solution is publicly available
online.\footnote{\url{https://git.persee.mines-paristech.fr/akylas.stratigakos/
ieee-cis-ppo}} For more details of this solution see Appendix G  in the
supplementary material.

\section{Discussion}
\label{sec:discussion}
A central aim of the competition was to design a problem in which both
forecasting and optimization are important tasks to perform well. By
establishing a well-defined, reproducible problem within a realistic setting, we
hope to encourage further methodological developments and comparative
evaluations in this domain. While this objective has been met to a significant
extent—evidenced by the fact that all but one shortlisted participant employed
competitive forecasting methodologies (with a Mean Absolute Scaled Error (MASE)
below 1)—the challenge of tightly integrating forecasting and optimization in
real-world decision-making remains complex. The competition successfully
demonstrated that neither component can be treated in isolation; however, the
nuanced interplay between probabilistic forecasting and optimization-based
decision-making still requires deeper exploration.

To the best of our knowledge, the number of time series used in this competition
was larger than in any similar undertaking before. However, it remains
relatively small, making it difficult to draw highly fine-grained conclusions.
Some models incorporated robustness by accounting for worst-case scenarios, but
given the limited number of time series and the short testing period, it is
unclear whether these strategies provided a tangible advantage. Robustness
effects are typically more evident over longer time horizons or across a broader
range of scenarios. Furthermore, no prior study has attempted a problem of this
scale, even for a single location, due to the inherent complexity involved.
Beyond battery scheduling, the competition also included a non-trivial lecture
scheduling component, adding layers of constraints and interdependencies that
significantly increase the problem’s difficulty. The decision to model and solve
this problem within a single university campus is grounded in practical
relevance—universities and institutions frequently operate microgrids with
centralized control, making them ideal testbeds for Predict+Optimize methods.
Moreover, extending the problem to multiple universities would not be
meaningful, as each institution has distinct operational constraints and
priorities. Rather than attempting to generalize across diverse energy systems
prematurely, this benchmark establishes an open-source reference for
state-of-the-art methods in both forecasting and optimization. By providing a
well-defined and reproducible problem, we aim to foster further advancements in
decision-focused learning and Predict+Optimize paradigms, encouraging deeper
exploration of the interplay between forecasting and optimization in real-world
applications.

The results of this competition highlight several important aspects of the
Predict+Optimize problem in renewable energy scheduling. All participants but
the 1st and 7th place solutions fed a single forecast into the optimizer, and
thus did not consider forecast uncertainty. The winning team employed stochastic
optimization to minimize the expected cost over a number of forecast scenarios.
The 7th team used a robust fix-and-optimize heuristic approach. While the method
succeeded in minimizing the worst case expected cost, the overall forecasting
performance was not as good as the other participants, which may have had an
effect on the optimization results. The competition results suggest that
inclusion of mitigating strategies for integration of uncertainty quantification
alone may not be sufficient to improve overall performance without a strong
foundation in forecasting quality. The advantages of integrating probabilistic
information have also been validated in related studies, such as the proceeds of
the Citylearn Challenge 2022 \cite{nweye2023citylearn}, where the winning team also
used a stochastic optimization approach to minimize the expected cost over a set
of forecast scenarios. The stochastic methods allow the optimization to hedge
against forecast errors by considering multiple scenarios, and thus provide a
way to improve the robustness of the solution. Given the recognition to the
potential gains from the use of stochastic optimization, it is important to note
that complexity and computational cost of stochastic methods can be significant,
making them less feasible for real-time applications without access to
high-performance computing resources. 

It is also worth noting that forecasting techniques used in the competition
among top-performers were mainly data-driven models, particularly tree-based
models. Most of the teams used a point forecast, and only two participants,
ranked 1st and 7th, incorporated some form of uncertainty quantification in
their forecasts. Both teams used a stochastic optimization approach over a set
of scenarios to minimize the expected cost. First-place winners (MA\&RE) used an
ensemble of LightGBM models with different granularity for weather features,
while the other team (AS) used a robust optimization to minimize the worst-case
expected cost, within the set of scenarios generated with a quantile regression
forest model based on the predictive density of a building with the highest
share of the total consumption. The rest of the participants used a single
forecast generated with gradient boosting machines, random forests, classical
time series models (ARIMA), and neural networks (ResNet). The competition
results suggest that the tree-based models were the most effective in
forecasting the energy demand and PV production. We observe that deep learning
models, such as ResNet, did not show a significant advantage over tree-based
models in this competition, and no transfer learning or pre-trained models were
used. We further note that the performance of transformer models in time series applications is a subject of ongoing debate in the forecasting community, as many proposals in this space have seriously flawed and narrow evaluations \cite{bergmeir2024fundamental,Hewamalage2023Forecast}. Participants not using such methods can be seen as an indication that these methods were not practical for our particular use case. Similarly, regarding traditional time series models, such as ETS or ARIMA, we hypothesize that participants found them to be not competitive. Our use case involves several time series with an opportunity to build a single model across all series, the problem has external
regressors, a multistep horizon in high resolution, and multiple seasonalities. Many traditional methods struggle with some of these characteristics, for example ETS and ARIMA only address single seasonalities of relatively short seasonal periods. 

There are discrepancies in the rankings of the methods
between forecasting and cost. While the MASE is a standard measure to evaluate
forecasts, the choice of this measure had certain implications for the
competition: The MASE weighs all series equally, while in terms of cost most
cost was concentrated in one time series (Building~3). Thus, for best
performance across both tasks, participants could have produced one forecast
that they used in the optimization, and another forecast to submit as their
forecast. This illustrates the challenges in the Predict+Optimize space. Other
error measures than the MASE would have likely led to different forecasting
methodologies, but presumably to similar overall outcomes in terms of energy
cost. The competition results showed a weak correlation between overall forecast
accuracy as evaluated in the competition, and optimization cost. Using a scaled
measure like MAE, and/or focusing on the time series with the largest values,
shows a higher correlation. 
Also, the participants did not find strong correlations during the competition
and one participant hypothesized that having a commercial solver such as Gurobi
and access to high-performance computing facilities were more important factors.
In contrast, the forecasting task could be performed on a single computer in
minutes.
Another participant noted that the validation data (Phase~1) included an
extremely large demand outlier, which affected the peak demand and the
respective peak tariff. In turn, this mitigated the impact of the objective
formulation (deterministic versus robust). Further, examining the results on
validation data (Phase~1) showed that, at least for the large instances, the
peak demand tariff comprised the biggest part of total energy cost. However, the
magnitude of the load to be scheduled during Phase~2 (relating to the respective
activities), was significantly smaller. If the problem instances are viewed as
data points from a problem distribution to be learned \cite{bengio2021machine},
this could be considered as a shift in the underlying distribution. Overall, the
peak tariff became less important during Phase~2, which somewhat obscured the
impact of forecast accuracy in total costs.

Thus, the competition results indicate that increased predictive accuracy does
not directly and not always translate into improved optimization performance,
and depends, among other things, on the forecast error measure used.

In a follow-up work, \citet{abolghasemi2022predict} further explored the
relationship between forecasting accuracy and optimization costs. The study
generated several scenarios, including consistent overforecast and underforecast
scenarios (perturbed), and computed their corresponding costs. The results
showed a Pearson correlation of 0.81 when using synthetic over-forecast and
under-forecast scenarios, and 0.9 for the competition participants' forecasts.
This indicates a strong association between forecast accuracy and optimization
costs. However, the study also found that this correlation is asymmetric,
meaning the impact of overforecasting and underforecasting is not the same.
Additionally, it suggests that any given forecast accuracy metric may not be the
most appropriate metric to minimize complex optimization costs.

The generalizability of these findings to broader contexts, such as other
microgrid systems, is promising. Most of the proposed methodologies and insights
from the competition apply to a broader class of resource-constrained scheduling
problems. Potential applications extend beyond energy systems to any domain
requiring integrated forecasting and optimization, such as supply chain
management, transportation planning, and financial portfolio optimization. As
indicated by top-performing solutions, MIP optimization paired with tree-based
forecasting algorithms shows superior performance. When formulation and
linearization of the constraints are challenging, heuristic-based evolutionary
algorithms become a viable alternative. While Model Predictive Control methods
are not directly applicable to the problem since they require a model of the
system dynamics instead of a fixed-time horizon, competition results reinforce
the idea that linear programming and mixed-integer programming are effective
tools for solving scheduling problems in renewable energy systems. 

The competition results revealed that none of the top-performing teams used
reinforcement learning (RL) for scheduling optimization, despite its success in
solving complex and dynamic problems in energy systems. While RL has
demonstrated potential in various applications, it did not provide a competitive
advantage in this setting, aligning with findings from similar challenges such
as the CityLearn Challenges in 2022 and 2023 \cite{nweye2023citylearn,
garmendia2024winning}, where the winning solutions ultimately relied on
classical optimization and heuristic methods rather than RL, known to be more
data-intensive \cite{nachum2018data}. We hypothize, that a key limitation of RL
in this competition was the absence of a dynamic simulation environment for
extensive training on real-world data, which is crucial for developing effective
RL policies. Given the constrained problem setup and limited dataset, RL
approaches likely fail to learn robust policies that could generalize
effectively beyond their training scope. These challenges underscore the
practical difficulties of applying RL in Predict+Optimize problems, particularly
when data availability is restricted and optimization constraints are highly
specific. The results suggest that, in such cases, well-designed classical
optimization techniques and heuristics remain more effective and reliable than
RL-based solutions. 

The competition results also highlight that significant performance gains come
from data cleaning. In general, most of the forecasting solutions involved
extensive manual data cleaning, such as outlier identification and removal,
which may be indicative of problems one would face in such a challenge
in the real world, where data quality issues are common. However, this makes
them less transferrable to an automated real-world production system. The
practical implications for real-time microgrid operations and scalability are
significant. The competition results suggest that integrating forecasting and
optimization can lead to substantial cost savings and more efficient energy
management. Compared to other studies, this competition stands out in
its focus on the integration of forecasting and optimization in a real-world
renewable energy scheduling problem. While previous studies have explored
similar problems, this competition explores the possibility for demand response
via timetable optimization. The open-sourced data and problem setting establish
a benchmark for future research using a real-world dataset.

\section{Conclusions}
\label{sec:conc}
This work has presented the results of the ``IEEE-CIS Technical Challenge on Predict+Optimize for Renewable Energy Scheduling,'' which was held to establish a benchmark dataset/problem, together with the state of the art in terms of performance on it, in a highly relevant research space that is currently lacking such a standard test bed. Out of 49 participants, the 7 shortlisted solutions have been presented here. Most top solutions converged to similar methodologies,
namely tree-based forecasting models and MIP optimization, with some notable
exceptions (one team used an evolutionary algorithm, another one a simple
heuristic for optimization, others used different forecasting methodologies).
The key contributions and findings of this study are summarized as follows:

\begin{itemize}
    \item Established a benchmark dataset/problem and evaluated the state of the art in the Predict+Optimize space for renewable energy scheduling.
    \item Demonstrated that tree-based forecasting models and MIP optimization are effective methodologies for this problem.
    \item Highlighted the importance of considering forecast uncertainty in optimization, as evidenced by the winning team's use of stochastic optimization.
    \item Identified the challenges of manual data cleaning and the need for
    automated solutions in real-world applications.
    \item Showed that increased predictive accuracy does not necessarily translate to improved optimization performance, also depending on the forecast error measure used.
    \item Suggested that future research should focus on developing better error measures, training models to directly minimize downstream optimization costs, and exploring other strategies in the Predict+Optimize space.
\end{itemize}

Quantitative results from the competition showed that the winning solution
achieved a significant reduction in energy costs compared to deterministic
approaches. Specifically, the 1st place solution achieved at least 2\% reduction
in energy costs compared to deterministic approaches. Furthermore, the
competition results highlight the potential misalignment between forecast
accuracy and downstream optimization performance, demonstrating that the most
accurate point forecast does not necessarily guarantee the best performance in
downstream optimization. 

Limitations of this study include the relatively small number of time series and
the short testing period, which may not fully capture the complexities of
real-world energy systems. Future research should explore the scalability of
these methods to larger and more complex systems. Furthermore, developing
multi-objective optimization frameworks to balance cost minimization with other
objectives, such as grid stability, battery health, or carbon footprint
reduction, could provide more holistic and sustainable solutions. Although the
computational costs are often infeasible for large-scale problems, integrated
gradient-based and gradient-free methods are promising directions for further
development.

\section*{Acknowledgment}

We are grateful to the IEEE Computational Intelligence Society to organize the
technical challenge series and to sponsor and assist with organization of this
edition in particular. We also thank the Department of Data Science and
Artificial Intelligence of Monash University for their sponsorship.
We would like to thank the late Prof Ariel Liebman (may he rest in peace, he is so dearly missed) for helpful discussions, and Prof Mark Wallace for forming part of the scientific committee. OikoLab kindly provided us ERA5 weather data. Finally,
we'd like to thank all other participants in the competition, in particular the
members of the QSZU-PolyU-Team, namely Qingling	Zhu, Minhui Dong, Wu Lin,
Yaojian Xu, Junchuang Cai, Junkai Ji, Qiuzhen Lin, and Kay Chen Tan.

\ifCLASSOPTIONcaptionsoff
  \newpage
\fi

\bibliographystyle{IEEEtranN}
\bibliography{IEEEabrv,Bibliography}

\begin{IEEEbiographynophoto}{Christoph Bergmeir}
Christoph Bergmeir is a María Zambrano (Senior) Fellow in the Department of Computer Science and Artificial Intelligence at University of Granada, Spain, and an Adjunct Senior Research Fellow in the Department of Data Science and Artificial Intelligence at Monash University. Before this, he was a Visiting Research Data Scientist at Meta Inc. (former Facebook Inc.) in California in the US, and a Senior Lecturer at Monash University. He has a track record of working in forecasting for capacity planning, sustainable energy, and supply chain, and has led teams that have delivered systems for short-term power production forecasting of wind and solar farms, and energy price forecasting. Christoph holds a PhD in Computer Science from the University of Granada, and an M.Sc. degree in Computer Science from the University of Ulm, Germany. He has over 10,000 citations and an h-index of 37. He has received more than AUD \$2.7 million in external research funding. Four of his publications on time series forecasting over the last years have been Clarivate Web of Science Highly Cited Papers (top 1\% of their research field).
\end{IEEEbiographynophoto}

\begin{IEEEbiographynophoto}{Frits de Nijs}
Frits de Nijs is a Research Fellow at Monash University within the Department of Data Science and Artificial Intelligence. His research focuses on multi-agent systems, reinforcement learning, sequential decision-making, demand response, and distributed energy resources. He has contributed significantly to the field through various projects and publications, including work on reinforcement learning for whole-building HVAC control and demand response.
Frits completed his PhD in Computer Science at the Delft University of Technology, where he specialized in multi-agent systems and sequential decision-making under uncertainty. His doctoral research involved developing algorithms for optimizing decision-making processes in stochastic environments with multiple actors.
\end{IEEEbiographynophoto}

\begin{IEEEbiographynophoto}{Abishek Sriramulu}
Abishek completed his PhD at Monash University, where he specialized in Graph Neural Networks and Time Series Forecasting. His doctoral research involved adaptive dependency learning in graph neural networks.
Prior to his PhD, Abishek worked as a Research Assistant at Monash University, contributing to various machine learning and AI projects. He also has industry experience as a Data Scientist at Servian, where he worked on data analytics and machine learning projects for various clients.
\end{IEEEbiographynophoto}

\begin{IEEEbiographynophoto}{Mahdi Abolghasemi}
Dr. Mahdi Abolghasemi is a Senior Lecturer in Data Science at Queensland University of Technology (QUT) and a researcher in forecasting, supply chain, and energy. With expertise in machine learning, statistical modeling, and decision-making, Mahdi has published over 30 articles in top-tier international journals and has collaborated with several industry partners to deliver forecasting solutions. He is a Chief Investigator at the Energy Transition Center at QUT, and his research is supported by funding from national and international organizations. He serves on the editorial board of the International Journal of Forecasting as a Book Editor and reviews for several other top journals in forecasting and data science.
\end{IEEEbiographynophoto}

\begin{IEEEbiographynophoto}{Richard Bean}
Dr Richard Bean is a Senior Energy Market Analyst at Energy Exemplar.
Previously, he worked at the University of Queensland’s Centre for Energy
Data Innovation. At the Centre, he analysed and modelled
low voltage electricity networks using near real-time big data. He worked closely with
centre partners Redback Technologies, Luceo, Energy Queensland,
and other entities to unlock new insights into electricity networks using IoT
technology at a neighbourhood level. Richard has extensive experience in
data science gained through working in academia in Australia and Iran,
for government (Queensland Health) and industry (ROAM Consulting and
the Australian Energy Market Operator). He has more than 60 publications
in areas including combinatorics, statistics, power systems, 
classical cryptography and active transport.
\end{IEEEbiographynophoto}

\begin{IEEEbiographynophoto}{John Betts}
Dr John Betts is a Senior Lecturer in the Department of Data Science and AI, Monash University. His research has applied computational modelling, optimisation, simulation and data analysis to investigate important societal problems across a diverse range of fields. These include models for renewable energy storage, inventory optimisation and cancer treatment. Ongoing work with social and political scientists investigates role of language and social dynamics in societal polarisation and the spread of social and political influence. 
\end{IEEEbiographynophoto}

\begin{IEEEbiographynophoto}{Quang Bui}
Quang Bui is a Research Fellow in the Department of Data Science and Artificial Intelligence at Monash University, where he has worked since September 2020. He focuses on advanced research in data science and artificial intelligence.
Before this role, Quang was a Research Assistant in Econometrics and Business Statistics and a Teaching Associate, contributing to both research and education at Monash University. He also worked as an Analyst and Risk Officer at LeasePlan, analyzing financial risks.
Quang holds a Bachelor of Commerce (Honours) in Econometrics and has taught various courses, consistently receiving high evaluations from students for his teaching excellence.
\end{IEEEbiographynophoto}

\begin{IEEEbiographynophoto}{Nam Trong Dinh}
Nam Trong Dinh is currently a PhD student in the School of Electrical and Electronic Engineering at the University of Adelaide, where he also received his Bachelor of Engineering and Honours degree in 2020. His research is funded by Watts A/S under the supervision of Dr Ali Pourmousavi, Prof Derek Abbott and Dr Mingyu Guo.
\end{IEEEbiographynophoto}

\begin{IEEEbiographynophoto}{Nils Einecke}
Nils Einecke received his M.Sc. degree (Diploma) in computer science, in 2007,
and his Ph.D. degree in engineering, in 2012, from the Technical University of
Ilmenau, Germany. He started working as a researcher at the Honda Research
Institute Europe in 2007. Since December 2022 he is leading the group of
Reliable Systems and Software at the Honda Research Institute Europe. His
current research interest includes computer vision, robotics, and energy
management.
\end{IEEEbiographynophoto}

\begin{IEEEbiographynophoto}{Rasul Esmaeilbeigi}
Rasul received his PhD in Mathematics from the University of Newcastle in 2020. In his PhD, he employed machine learning, dynamic programming, mixed integer linear programming and two-stage stochastic programming approaches to solve variants of a supply chain network design problem. He developed decomposition algorithms that scale well with the number of uncertain scenarios, conducted extensive benchmarking under various scenarios, and provided managerial insights for a real case study. In his post-doctoral role at Deakin University, Rasul applied his expertise to develop a decision support system (called optimiser engine) for a real-world resource planning problem. An extensive simulation study demonstrated that the proposed optimiser significantly improves the efficiency of client’s operations over time, thereby potentially saving millions of dollars for them. Due to the significance of the theoretical and practical contributions, this work was selected as a semi-finalist of INFORMS Franz Edelman Award (2021).
\end{IEEEbiographynophoto}

\begin{IEEEbiographynophoto}{Scott Ferraro}
Scott was part of the leadership team -- including a period as acting CEO -- at ClimateWorks Australia. There, he co-led the 'Pathways to Deep Decarbonization in 2050' initiative, translating net-zero strategies for government bodies including international, Federal, State and local governments, and corporate partners such as Vicinity Centres, Melbourne Water, Bunnings, and Monash University. He then went on to lead Monash University's \$135m Net Zero Initiative integrating operational deployment with research, education, and industry collaboration. Scott led the development of the Monash Microgrid, facilitating building and grid interactivity, and fostering collaboration across universities, government, and industry, including the establishment of the Monash-Engie Net Zero Solutions Alliance. On top of this work, he is currently a board member of the Energy Efficiency Council, co-led the establishment of the Electric Vehicle Council, and Chaired the Surf Coast Renewable Energy Taskforce. Coupled with his professional experience, as a qualified Environmental Engineer with a Masters degree in Corporate Environmental and Sustainability Management, Scott has a deep understanding of the technical and commercial challenges and solutions to deliver Net Zero for businesses.
\end{IEEEbiographynophoto}

\begin{IEEEbiographynophoto}{Priya Galketiya}
Priya is an energy tech specialist spending the last 5 years working on C\&I energy projects with Monash University, n0de and Green Transition Co focusing on microgrids and firming solutions.  He holds an Electrical Engineering degree from the University of Queensland.
\end{IEEEbiographynophoto}

\begin{IEEEbiographynophoto}{Evgenii	Genov}
Evgenii Genov is a PhD researcher at the EVERGI group, Vrije Universiteit Brussel. He studies the applications of deep learning with energy forecasting.
Evgeny has obtained a double MSc degree in energy engineering from KU Leuven and KTH, Stockholm. His MSc thesis is a study of digital filter design for fault detection in high-voltage DC grid power systems. He completed BSc in Utrecht University, majored in Physics and Mathematics. 
\end{IEEEbiographynophoto}

\begin{IEEEbiographynophoto}{Robert Glasgow}
Robert is a skilled software architect with extensive experience designing and delivering innovative software solutions. Prior to joining Monash University in 2018, he spent the majority of his career in the energy sector where he held positions from software engineer to company director. While based in the Monash eResearch Centre, he led the NetZero TEM \& IoT Platform R\&D. Having returned to an industry role, Robert is now a software architect with EDMI where he is working on large scale energy solutions.
\end{IEEEbiographynophoto}

\begin{IEEEbiographynophoto}{Rakshitha Godahewa}
Rakshitha Godahewa is a PhD Researcher in the Department of Data Science and Artificial Intelligence at Monash University. Rakshitha holds a B.Sc. degree in Computer Science and Engineering from the University of Moratuwa, Sri Lanka.
\end{IEEEbiographynophoto}

\begin{IEEEbiographynophoto}{Yanfei Kang}
Yanfei Kang is an Associate Professor at the School of Economics and Management of Beihang University and Head of the Department of Quantitative Economics and Business Statistics. She received her PhD degree from Monash University and previously worked as a postdoctoral researcher at Monash University and Baidu Inc.’s Big Data Group as a senior R\&D developer. Her research is published in various academic journals such as European Journal of Operational Research, International Journal of Forecasting, International Journal of Production Research, Statistical Analysis and Data Mining, Machine Learning, Pattern Recognition, among others.
\end{IEEEbiographynophoto}

\begin{IEEEbiographynophoto}{Steffen	Limmer}
Steffen Limmer received the M.Sc.(Diploma) in computer science from the University of Jena, Germany in 2009. In 2016, he received the Ph.D. degree in engineering from
the University of Erlangen-Nürnberg, Germany, where he worked from 2009 to 2016 as scientific assistant at the chair of computer architecture.
Since December 2016, Dr. Limmer is senior scientist at the Honda Research Institute Europe. His current research topics are optimization and data-driven modeling in the context of energy management systems.
\end{IEEEbiographynophoto}

\begin{IEEEbiographynophoto}{Luis	Magdalena}
Luis Magdalena is Full Professor of Computer Science and Artificial Intelligence at Universidad Politecnica de Madrid, Spain. He is co-author of the books ``Explainable Fuzzy Systems - Paving the Way from Interpretable Fuzzy Systems to Explainable AI Systems'' (Springer. 2021) and ``Genetic Fuzzy Systems: Evolutionary Tuning and Learning of Fuzzy Knowledge Bases'' (World Scientific, 2001), and has authored more than 150 papers in journals and conferences. He has been involved in over fifty research projects mostly related to computational intelligence, robotics and automation. Prof. Magdalena is President-elect of the IEEE Computational Intelligence Society and is also Vice-president of the Spanish Society for Artificial Intelligence. 
\end{IEEEbiographynophoto}

\begin{IEEEbiographynophoto}{Pablo Montero-Manso}
Pablo Montero-Manso is a Lecturer in the Discipline of Business Analytics, University of Sydney. His research interests focus on Machine Learning and Statistics for Time Series Analysis: clustering, classification and forecasting.
Pablo has developed several successful methodologies for automatic forecasting on large datasets. In 2018, he co-developed an award-winning forecasting methodology and during 2020 and 2021 his research was applied to the forecasting of the COVID-19 pandemic in both Australia and Spain. He has authored and maintains multiple open source data analysis tools. His research and software tools are being used in academia, industry, and the public sector.
Pablo received his PhD from the University of A Coruna, Spain. Before joining the University of Sydney, he was a post-doctoral fellow at Monash University.
\end{IEEEbiographynophoto}

\begin{IEEEbiographynophoto}{Daniel Peralta}
Dr. Daniel Peralta is an assistant professor at the Department of Computer Science and Artificial Intelligence of the University of Granada (Spain). He obtained his PhD at the same university in 2016, tackling large-scale fingerprint identification. His research has focused on machine learning, especially in large-scale scenarios, and has involved several collaborations with industry to apply such techniques on problems ranging from railway maintenance scheduling to compound activity prediction. Between 2016 and 2024, Dr. Peralta carried out several post-doctoral periods at Ghent University (jointly with VIB and imec), and applied his research on biological data and wireless signal data analysis.  
\end{IEEEbiographynophoto}

\begin{IEEEbiographynophoto}{Yogesh	Pipada Sunil Kumar}
Yogesh	Pipada Sunil Kumar is with the School of Electrical and Electronics Engineering, University of Adelaide, Adelaide, Australia. He is currently an Industrial PhD candidate with a Masters in Engineering (Electrical) from the University of Melbourne. He also has some former work experience in solar farm and capacitor bank commissioning.
His current research is on optimal bidding strategies for Aggregators of Distributed Energy Resources, in the Wholesale and Real time energy markets.
\end{IEEEbiographynophoto}

\begin{IEEEbiographynophoto}{Alejandro Rosales-P\'erez}
Alejandro Rosales-Pérez received the B.S. degree in Electronic Engineering from the Instituto Tecnológico de Tuxtla Gutiérrez, Chiapas, Mexico, in 2008, and the M.Sc. and Ph.D. degrees in Computer Science from the Instituto Nacional de Astrofísica, Óptica y Electrónica, Puebla, Mexico, in 2011 and 2016, respectively. Since 2020, Dr. Rosales-Pérez is with Centro de Investigación en Matemáticas, A.C. (CIMAT), where he is currently a researcher associate C.
Dr. Rosales-Pérez was a recipient of the Award by the National Association of Education Institutions in Information Technology in 2016 for his Doctoral Thesis. His thesis was the first runner-up in National Contest on Artificial Intelligence, granted by the Mexican Society on Artificial Intelligence in 2016. Since 2017, Dr. Rosales-Pérez is a member of the Mexican Academy for Computing. He has been recognized by the Mexican National System of Researchers with Level I, 2017-2019, supported by the Mexican National Council for Science and Technology.
\end{IEEEbiographynophoto}

\begin{IEEEbiographynophoto}{Julian	Ruddick}
Julian Ruddick is a PhD student at the Vrije Universiteit Brussel, MOBI. His PhD thesis focuses on applying artificial intelligence to energy management problems.
In particular, he implements and compares different methods to improve the efficiency of energy management systems in microgrids. He is currently working on the MAMûET project.
Julian obtained his Master's Degree in Computer Science Engineering at the Université Libre de Bruxelles (ULB) in 2019, with a specialisation in artificial intelligence. His master thesis is a comparison of different artificial intelligence based design methods for swarm robotics.
\end{IEEEbiographynophoto}

\begin{IEEEbiographynophoto}{Akylas	Stratigakos}
Akylas Stratigakos received his Diploma in Electrical and Computer Engineering from the University of Patras, Greece, in 2016. Currently, he is pursuing his PhD degree at Mines Paris, PSL University, focusing on forecasting, decision-making under uncertainty, and machine learning applications in power systems.
\end{IEEEbiographynophoto}

\begin{IEEEbiographynophoto}{Peter	Stuckey}
Professor Peter J. Stuckey is a Professor in the Faculty of Information Technology at Monash University, and project leader in the Data61 CSIRO laboratory. Peter Stuckey is a pioneer in constraint programming, the science of modelling and solving complex combinatorial problems. 
His research interests include: discrete optimization; programming languages, in particular declarative programming languages;  constraint solving algorithms; bioinformatics; and constraint-based graphics.
He enjoys problem-solving in any area, having publications in e.g. databases, timetabling, and system security, and working with companies such as Oracle and Rio Tinto on problems that interest them.
Peter Stuckey received a B.Sc and Ph.D both in Computer Science from Monash University in 1985 and 1988 respectively.
In 2009 he was recognized as an ACM Distinguished Scientist. In 2010 he was awarded the Google Australia Eureka Prize for Innovation in Computer Science for his work on lazy clause generation. He was awarded the 2010 University of Melbourne Woodward Medal for most outstanding publication in Science and Technology across the university. In 2019 he was elected as a Fellow of the Association for the Advancement of Artificial Intelligence.
\end{IEEEbiographynophoto}

\begin{IEEEbiographynophoto}{Guido Tack}
Guido Tack is an Associate Professor in the Department of Data Science and Artificial Intelligence, at the Faculty of Information Technology, Monash University.
His research focuses on combinatorial optimization, in particular architecture and implementation techniques for constraint solvers, translation of constraint modelling languages, and industrial applications. Guido leads the development of the MiniZinc constraint modelling language and toolchain. He is one of the main developers of Gecode, a state-of-the-art constraint programming library.
Guido's broader research interests include programming languages and computational logic.
Guido graduated and received his doctoral degree (Dr.-Ing.) from the Department of Computer Science, Saarland University, Germany. Before joining Monash University as a Lecturer and Monash Larkins Fellow in February 2012, he worked as a post-doctoral researcher at NICTA Victoria Laboratory, Saarland University (Germany), and K.U. Leuven (Belgium).
\end{IEEEbiographynophoto}

\begin{IEEEbiographynophoto}{Isaac Triguero}
Isaac Triguero received his M.Sc. and Ph.D. degrees in Computer Science from the University of Granada, Granada, Spain, in 2009 and 2014, respectively. He is a currently enjoying a Distinguished Senior Research Fellowship at the University of Granada. His work is mostly concerned with the research of novel methodologies for big data analytics. Dr Triguero has published more than 90 international publications in the fields of Big Data, Machine Learning and Optimisation (H-index=35 and more than 5700 citations on Google Scholar). He is a Section Editor-in-Chief of the Machine Learning and Knowledge Extraction journal, and an associate editor of the Big Data and Cognitive Computing journal and the IEEE Access journal. He has acted as Program Co-Chair of the IEEE Conference on Smart Data (2016), the IEEE Conference on Big Data Science and Engineering (2017), and the IEEE International Congress on Big Data (2018). Dr Triguero is currently co-leading  two projects on General Purpose Artificial Intelligence:  a €1.6M University-Industry Research Grant funded by the European Union-Next Generation EU as well as a €120K Knowledge Generation Project, funded by the Ministry of Science, Innovation and Universities of Spain. 
\end{IEEEbiographynophoto}

\begin{IEEEbiographynophoto}{Rui	Yuan}
Rui	Yuan is with the School of Electrical and Electronics Engineering, University of Adelaide, Adelaide, Australia. As a PhD candidate working on an industrial program, his research is aiming to develop, test, and verify statistical models based on machine learning techniques to quantify prosumers’ responsiveness to time-varying prices in real-time. His research interest includes Power and Energy System Engineering, Pattern Recognition, Data Mining, and Explainable Machine Learning in Power System. His other skills include forecasting and modelling (achieved 5th place in IEEE-CIS DATA CHALLENGE 2021), web development (served as web chair of the 32nd Australasian Universities Power Engineering Conference), and synthetic data generation (Synthetic dataset published in Scientific Data by Nature).
\end{IEEEbiographynophoto}

\EOD

\end{document}